\documentclass{article} 
\usepackage{iclr2027_conference,times}

\usepackage{amsmath,amsfonts,bm}

\def\eqref#1{equation~\ref{#1}}

\def\1{\bm{1}}

\DeclareMathAlphabet{\mathsfit}{\encodingdefault}{\sfdefault}{m}{sl}
\SetMathAlphabet{\mathsfit}{bold}{\encodingdefault}{\sfdefault}{bx}{n}

\usepackage{enumitem}

\usepackage{multicol}

\renewcommand{\paragraph}[1]{\vspace{0em}\noindent\textbf{#1}.}

\definecolor{turquoise}{cmyk}{0.65,0,0.1,0.3}
\definecolor{purple}{rgb}{0.65,0,0.65}
\definecolor{dark_green}{rgb}{0, 0.5, 0}
\definecolor{orange}{rgb}{0.8, 0.6, 0.2}
\definecolor{red}{rgb}{0.8, 0.2, 0.2}
\definecolor{darkred}{rgb}{0.6, 0.1, 0.05}
\definecolor{blueish}{rgb}{0.0, 0.3, .6}
\definecolor{light_gray}{rgb}{0.7, 0.7, .7}
\definecolor{pink}{rgb}{1, 0, 1}
\definecolor{greyblue}{rgb}{0.25, 0.25, 1}

\usepackage{url}
\usepackage{booktabs}

\usepackage[letterpaper,lmargin=1.5in,rmargin=1.5in,tmargin=1in,bmargin=1in]{geometry}
\usepackage[utf8]{inputenc}
\usepackage[T1]{fontenc}
\usepackage{amsmath}
\usepackage{amsthm}
\usepackage{amsfonts}
\usepackage{amssymb}
\usepackage{microtype}
\usepackage{graphicx}
\usepackage{subcaption}
\usepackage{booktabs}
\usepackage{nicefrac}
\usepackage{xcolor}
\usepackage{float}
\usepackage[skins,listings,breakable]{tcolorbox}
\usepackage{adjustbox}
\usepackage[pagebackref=true,colorlinks,citecolor=brown]{hyperref}
\usepackage[font=small]{caption}

\usepackage{etoc}           
\etocsettagdepth{mainpaper}{none}
\etocsettagdepth{appendix}{subsection}
\usepackage{amsmath,amssymb,amsthm,mathtools}
\usepackage{enumitem}
\usepackage{hyperref}
\usepackage{cleveref}

\usepackage{xcolor}

\usepackage{graphicx} 
\usepackage{wrapfig}
\usepackage{multirow}

\usepackage{algorithm}
\usepackage{algpseudocode}
\usepackage{natbib}

\usepackage{listings}
\usepackage{booktabs}
\usepackage{xcolor}

\lstdefinestyle{prompt}{
basicstyle=\ttfamily\footnotesize,
breaklines=true,
breakatwhitespace=true,
columns=fullflexible,
keepspaces=true,
frame=single,
framerule=0.4pt,
xleftmargin=6pt,
xrightmargin=6pt,
aboveskip=6pt,
belowskip=6pt,
}

\algnewcommand\algorithmicwrite{\textbf{write}}
\algnewcommand\Write{\item[\algorithmicwrite]}
\algnewcommand\Note[1]{\State \textbf{note} #1}
\algnewcommand\Log[1]{\State \textbf{log} #1}

\usepackage{caption}
\usepackage{cleveref}
\crefname{figure}{Fig.}{Figs.}
\Crefname{figure}{Fig.}{Figs.}

\tcbuselibrary{listings,skins}
\usepackage{xcolor}

\usepackage{booktabs}
\usepackage{array}
\usepackage{tabularx}
\usepackage{ makecell, pifont, graphicx}

\definecolor{codegray}{RGB}{245,245,245}
\definecolor{codegreen}{RGB}{0,120,80}
\definecolor{codeblue}{RGB}{40,80,160}

\lstdefinestyle{pythonstyle}{
    language=Python,
    backgroundcolor=\color{codegray},
    basicstyle=\ttfamily\footnotesize,
    keywordstyle=\color{codeblue},
    commentstyle=\color{codegreen},
    stringstyle=\color{purple},
    numbers=left,
    numberstyle=\tiny\color{gray},
    numbersep=8pt,
    breaklines=true,
    showstringspaces=false,
    frame=single,
    framesep=5pt,
    tabsize=4,
    columns=fullflexible
}

\newtcolorbox[auto counter, crefname={Box}{Boxes}]{promptbox}[2][]{
    enhanced,
    breakable,
    colback=gray!3,
    colframe=gray!50,
    boxrule=0.5pt,
    arc=2mm,
    left=3mm, right=3mm, top=2mm, bottom=2mm,
    title={Box~\thetcbcounter: #2},
    fonttitle=\bfseries,
    coltitle=black,
    colbacktitle=gray!15,
    titlerule=0.5pt,
    toptitle=1mm, bottomtitle=1mm,
    #1
}

\newtcblisting[auto counter, number within=section]{myskillbox}[3][]{
    breakable,            
    enhanced,             
    title={%
      \begin{minipage}[c]{0.09\linewidth}
        \includegraphics[width=\linewidth]{#3}%
      \end{minipage}%
      \hfill
      \begin{minipage}[c]{0.85\linewidth}
        \textbf{Skill \thetcbcounter:} #2
      \end{minipage}%
    },
    label={#1},
    colback=gray!5,
    colframe=black,
    boxrule=0.6pt,
    arc=0mm,
    top=0pt, bottom=6pt, left=6pt, right=6pt,
    listing only,
    listing options={
        style=pythonstyle,
        backgroundcolor={},
        frame=none,
        numbers=none,
        escapeinside={(*@}{@*)}
    }
}

\title{Multimodal Filesystem Memory for Long-Horizon Embodied Agents}

\newcommand{\harvard}{1}
\newcommand{\kempner}{2}
\newcommand{\nyu}{3}
\newcommand{\uiuc}{4}
\newcommand{\neu}{5}
\newcommand{\massit}{6}

\author{Chloe Su$^{\harvard,\kempner,\star}$ \quad
Jiahan Li$^{\nyu,\star}$ \quad
Zhenting Qi$^{\harvard}$ \quad
Chaoran Cheng$^{\uiuc}$ \quad
Haoyu Chen$^{\harvard}$\\
\bf Sejoon Jun$^{\neu}$ \quad
Weirui Ye$^{\massit}$ \quad
Yilun Du$^{\harvard,\kempner}$ \\
\\
$^{\harvard}$Harvard \qquad
$^{\kempner}$Kempner Institute \qquad
$^{\nyu}$NYU \qquad
$^{\uiuc}$UIUC \qquad
$^{\neu}$NEU \qquad
$^{\massit}$MIT
\vspace{0.7em}
\\
\href{https://github.com/Hither1/MFM}{%
    \raisebox{-0.12em}{\includegraphics[height=1.05em]{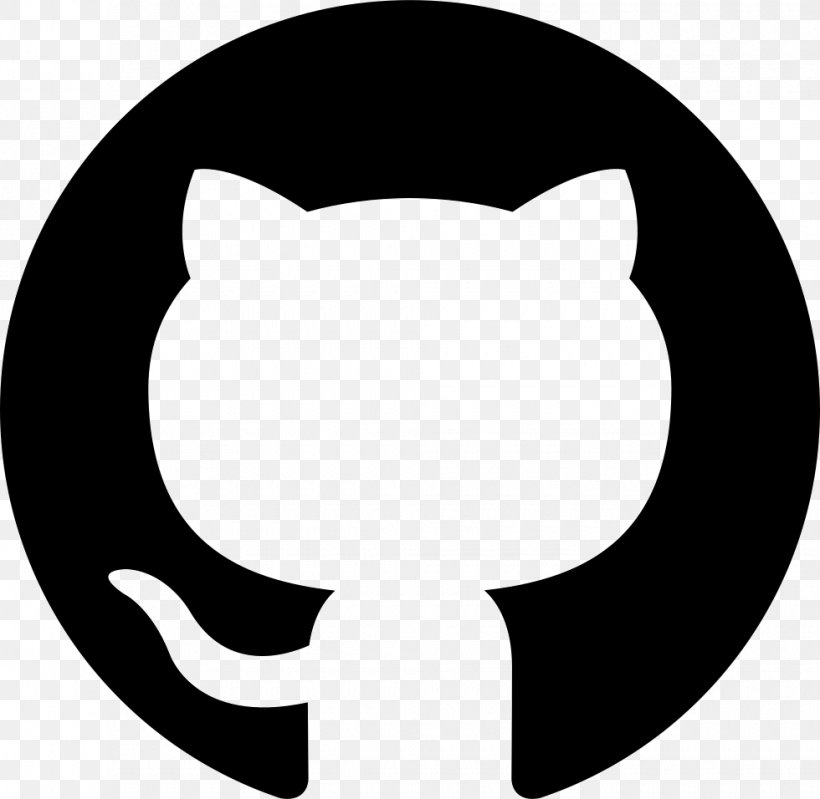}}~\textbf{GitHub}%
}
\qquad
\href{https://hither1.github.io/MFM-project/}{%
    \raisebox{-0.12em}{\includegraphics[height=1.05em]{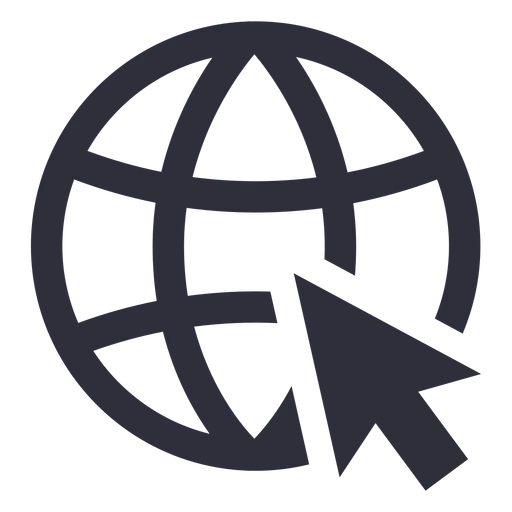}}~\textbf{Project Page}%
}
}

\iclrfinalcopy 
\begin{document}

\maketitle

\ificlrfinal
\begingroup
\renewcommand{\thefootnote}{$\star$}
\footnotetext{Equal contribution.}
\endgroup
\fi

\lhead{Preprint}
\begin{abstract}

Agents operating over long horizons need an effective memory that allows them to reuse their experience throughout long interactions. If the entire history of an interaction is retained, it becomes difficult to find the relevant experience, while compressing it into summaries may result in the loss of visual and spatial details that are required later on. We present \textbf{MFM}, a multimodal filesystem memory architecture that keeps complementary types of agent memory. \textbf{Episodic memory} stores the observations, actions, and outcomes, and \textbf{spatial memory} associates experiences with locations. \textbf{Semantic and procedural memory} build up general knowledge and reusable methods of acting, and \textbf{working memory} holds the current plans. MFM makes use of searchable textual records to connect these different representations to the underlying multimodal episodes. The agents are able to find previous experiences by means of contextual cues and to examine the original observations in order to recover details that are missing from their textual descriptions. Through consolidation, the agents convert experience into higher-level knowledge and procedures while still keeping references to the original episodes. In the case of long-horizon Minecraft and computer-use tasks, MFM leads to improved performance with three base models. With GPT-6-Astra, MFM completes \emph{Mine Diamond from Scratch} and reaches 16 out of 24 milestones on the \emph{Ender Dragon} task, as against 12.7 for the best baseline. On CUABench-KiCad it attains a mean reward of 0.440, as compared to 0.400 for the best baseline and 0.267 in the case of no persistent memory.


\end{abstract}

\etocdepthtag.toc{mainpaper}
\section{Introduction}

Agents that are intelligent and capable of operating autonomously over long periods in complicated, open-ended environments need to remember what they have seen, discovered, and previously done~\citep{park2025mrsteve,wang2026nativemem}. A large part of this kind of memory will generally be visual or spatial in character, for instance the route by which one can get back to an object that it has seen before. It is not clear what the appropriate form of compact memory the agent should use to store such observations. If one stores everything directly then it becomes difficult to find relevant experiences as the amount of history increases. Conversely, summaries of the memory enable quicker retrieval, but may result in important details being discarded later \citep{lu2026mma,zou2026task}. 

\looseness=-1
We represent memory as a multimodal filesystem (MFM) which simultaneously holds different types of memory. We organize the memory based on the distinctions between various human memory systems~\citep{tulving1972episodic,squire2004memory}. In the MFM, episodic memory preserves past observations and interactions, spatial memory connects experiences to locations in the environment, semantic memory records facts and relationships, procedural memory contains reusable methods of acting, and working memory holds the current plan and the immediate subgoals. The five memory functions are realised using three common storage elements: episodic records, an associative index, and agent-written memory. These elements are arranged as folders and files in a shared filesystem, with references that link knowledge, procedures, and plans to their supporting episodes.

Previous work that has used a filesystem to store agent memory has shown the effectiveness of searchable textual records of interaction~\citep{fox2026pro}. With embodied agents, textual descriptions can fail to include important visual or spatial information. VISTA~\citep{vista2026} keeps visual observations and allows frames to be revisited using a turn or frame identifier, thus holding on to information that goes beyond its textual summaries. Yet it can be difficult to find the relevant experience, particularly as the memory increases. MFM offers an explicit structuring of the memory in the form of a filesystem, which enables quick retrieval of context and the aggregation of skills and knowledge.


\begin{figure*}[ht]
  \centering
    \centering
    \includegraphics[width=\linewidth]{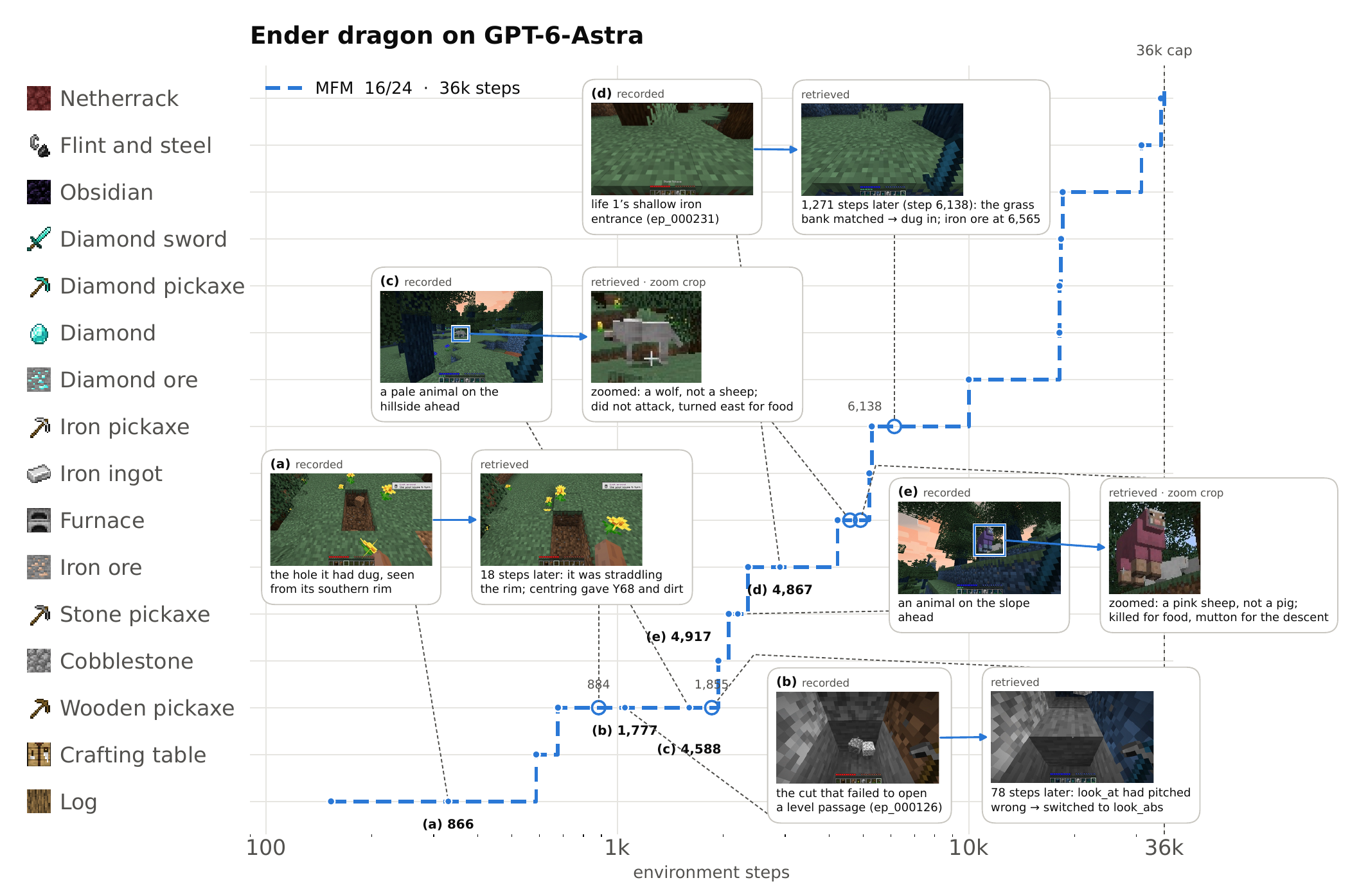} 
    \label{fig:}
    \vskip -0.5cm
  \caption{\small \textbf{MFM's Memory Evolution in Minecraft.} Over 36,000 steps, MFM continuously updates its memory and retrieves from it to support decision-making.
}
  \label{fig:teaser}
  \vspace{-10pt}
\end{figure*}

As shown in \Cref{fig:overview}, \textbf{associative indexing} links these representations via searchable cues to the stored evidence. Text acts as an \emph{addressing layer} for multimodal records: the agent begins by searching the descriptions of events, actions, and locations and then looks up the references in order to examine the corresponding observations. For instance, it can find a previous visit to a cave by means of its coordinates or by referring to an action that was attempted, check the recorded view, and then compare it with the current observation. The cues employed to locate an episode need not include a description of all the details of the episode. This enables us to navigate a long history of past episodes using compact records and at the same time retain detailed visual information which may only become relevant later.

With increasing experience, MFM employs self-evolving consolidation in order to update its knowledge, skills, and current working plans. Specific observations can lead to the establishment of new facts or relationships, useful action sequences can be turned into reusable procedures, and new information discovered can cause the agent to alter its present plan. These updates link episodic experience to semantic, procedural, and working memory. For instance, the agent might connect a resource with a cave that has just been explored, record a reliable procedure for getting to it, and change its route if a subsequent observation indicates that the entrance is blocked. Since the memories written by the agent include references to the episodes that support them, the agent can go back and look at the original evidence whenever its interpretation becomes uncertain or is contradicted. The evolution of the memory is illustrated in \Cref{fig:teaser}.

Our work consists of three parts. Firstly, we propose a method for organising a multimodal agent's memory within a single filesystem, covering episodic, spatial, semantic, procedural, and working memory. Secondly, we implement this organisation in MFM, which combines cue-based retrieval and visual inspection with consolidation into revisable knowledge and reusable procedures. Thirdly, we demonstrate through long-horizon Minecraft and computer-use tasks that MFM surpasses both memory-free and memory-augmented baseline methods when using three base models, and that it is able to reuse its memory across different tasks. Using GPT-6-Astra, our system manages to complete the MCU \emph{Mine Diamond from Scratch} task in all instances and achieves 16 out of 24 milestones on \emph{Ender Dragon}, whereas the best baseline achieves an average of 12.7. On CUABench-KiCad, its mean reward is 0.440, compared to 0.400 for the best baseline and 0.267 for an agent lacking persistent memory.

\vspace{-10pt}
\section{Related Work}
\vspace{-5pt}
\label{sec:rw}

\begin{table}[t]
\centering
\small
\caption{\small \textbf{MFM versus prior agent memory systems} across key memory properties.}
\label{tab:positioning}
\resizebox{\linewidth}{!}{%
\begin{tabular}{lcccccl}
\toprule
Method & \makecell{Raw multimodal\\evidence} & \makecell{Index with\\evidence pointers} & \makecell{Spatial\\memory} & \makecell{Agent-writable\\memory} & \makecell{Knowledge\\linked to evidence} & Retrieval \\
\midrule
Voyager \citep{wang2024voyager}    & \ding{55} & \ding{55} & \ding{55} & \ding{51}$^{a}$ & \ding{55} & Embedding similarity \\
JARVIS-1 \citep{wang2025jarvis}    & \ding{51}$^{b}$ & \ding{55} & \ding{55} & \ding{51}$^{b}$ & \ding{55} & Multimodal similarity \\
MrSteve \citep{park2025mrsteve}    & \ding{51} & \ding{55} & \ding{51} & \ding{55} & \ding{55} & Place/event lookup \\
PRO-LONG \citep{fox2026pro}        & \ding{55}$^{c}$ & \ding{55} & \ding{55} & \ding{51}$^{d}$ & \ding{55}$^{d}$ & Regex / code over log \\
VISTA \citep{vista2026}            & \ding{51} & \ding{55}$^{e}$ & \ding{55} & \ding{51}$^{e}$ & \ding{55} & Inspect by frame ID \\
\midrule
\textbf{MFM (ours)} & \ding{51} & \ding{51} & \ding{51} & \ding{51} & \ding{51} & Cue query + evidence inspection \\
\bottomrule
\multicolumn{7}{l}{\scriptsize $^{a}$Skill library (code) only. \; $^{b}$Successful trajectories only. \; $^{c}$Text log only. \; $^{d}$Optional notes, not linked to log entries. \; $^{e}$Frame IDs without a searchable index; free-form notes.} \\
\end{tabular}}
\vspace{-15pt}
\end{table}

\textbf{Filesystem and Multimodal Memory.}
Approaches that use a filesystem allow agents to search through and modify persistent records. PRO-LONG has demonstrated that searchable textual interaction logs can support long-horizon reasoning \citep{fox2026pro}, and in other cases filesystems are used as organizational structures for agent memory \citep{zhou2026filesystem}. Multimodal memory systems are designed to organize long sequences of visual data in order to enable question answering and reasoning \citep{fan2024videoagent,liu2026longvideoagent,long2026seeing}. Other research looks into task-conditioned retention \citep{zou2026task}, retrieval reliability in the face of temporal decay and conflicting evidence \citep{lu2026mma}, query-aware relational organization \citep{li2026graphmemix}, and analytic operations performed on accumulated observations \citep{tian2026beyond}. Table~\ref{tab:positioning} summarizes the differences between MFM and related systems.

\textbf{Memory for interactive agents with a long horizon.}
The ability to carry out long-horizon interactions demands a type of memory that enables continuous perception, exploration, and planning. The latest ARC-AGI-3 systems integrate foundation models with interaction harnesses and external memory \citep{arcagi3,fox2026pro,schema2026,rodionov2026executable}, and VISTA keeps records of visual information that agents can go back to and compare \citep{vista2026}. In the field of robotics, efforts are directed towards multi-scale memory, the learning of history compression, and the development of benchmarks for memory-dependent policies \citep{torne2026mem,wang2026nativemem,dai2026robomme}. Spatial memory also helps with navigation and reasoning \citep{ge2026evomemnav,zhang2026spatial}. In Minecraft, Voyager creates a collection of executable skills \citep{wang2024voyager}, JARVIS-1 retrieves successful plans from its multimodal memory \citep{wang2025jarvis}, and later agents introduce hybrid multimodal memory \citep{li2024optimus}, what-where-when representations \citep{park2025mrsteve}, and accumulated knowledge as a means of self-improvement \citep{xie2026mineevolve}. New benchmarks have recently emphasised exploration and long-horizon competence \citep{zheng2025mcu,ju2026mineexplorer,zhou2026can}. Although PRO-LONG indexes only text and VISTA revisits frames without having a searchable index, MFM provides a general-purpose multimodal episodic filesystem.

\vspace{-5pt}
\section{Multimodal Filesystem Memory}
\vspace{-5pt}
\label{sec:method}

\begin{figure*}[ht]
  \centering
    \centering
    \includegraphics[width=\linewidth]{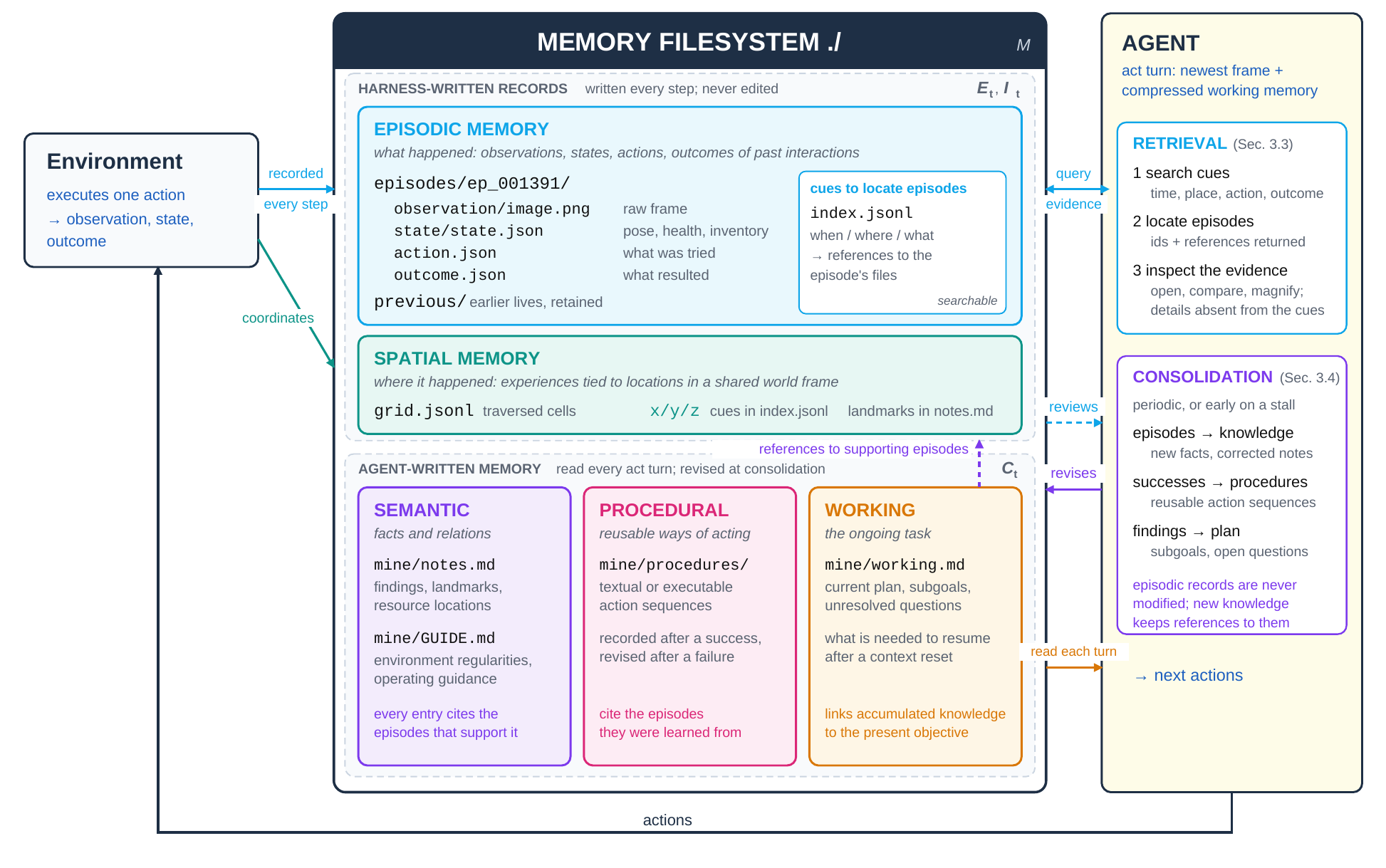} 
    \label{fig:}
    \vspace{-20pt}
  \caption{\small \textbf{Overview of MFM.} Five forms of memory in one filesystem. Harness-written \textbf{episodic memory} records observations, states, actions, and outcomes every step, with searchable cues for locating them; \textbf{spatial
memory} ties experiences to locations. Agent-written \textbf{semantic}, \textbf{procedural}, and \textbf{working memory} hold facts, reusable procedures, and the current plan, each citing its supporting episodes. The agent retrieves by
searching cues and inspecting original evidence (Sec.~\ref{sec:retrieval}), and consolidates experience into knowledge, procedures, and plans without modifying episodic records (Sec.~\ref{sec:consolidation}).
}
  \label{fig:overview}
  \vspace{-10pt}
\end{figure*}

We present \textbf{MFM}, a multimodal filesystem memory architecture designed for long-horizon agents. This architecture incorporates three complementary elements: \textbf{episodic records} which store the multimodal experience, a \textbf{searchable index} that enables textual access to these records, and \textbf{agent-written memory} which structures plans, findings, environmental knowledge, and reusable procedures. \Cref{fig:overview} shows the structure of this approach, while \Cref{fig:memory-detail} illustrates an example from Minecraft, covering the processes of episodic recording and indexing as well as retrieval and consolidation. \Cref{sec:representation} explains the various memory components and the way in which searchable records are linked to the multimodal evidence, \Cref{sec:retrieval} describes how the agent finds and examines relevant episodes, and \Cref{sec:consolidation} outlines how the accumulated experience is consolidated into reusable knowledge without losing the original evidence. \Cref{alg:mfm-main} gives a summary of the entire loop.

\subsection{Problem Setup}
\label{sec:problem-setup}
An agent interacts with an environment over a long horizon to achieve a task objective $g$. At step $t$, it receives an observation $o_t=(v_t,x_t)$, where $v_t$ is a visual observation and $x_t$ contains textual or structured information, and selects an action $a_t$. Its interaction history is $H_t=(o_1,a_1,\ldots,a_{t-1},o_t)$. As this history grows beyond the model's context budget, the agent must retain information outside its active context and retrieve it when needed. We represent this external memory as $M_t=(E_t,I_t,C_t)$, where $E_t=(e_1,\ldots,e_{N_t})$ stores the $N_t$ completed episodes retained at step $t$, $I_t$ provides an index for locating these episodes, and $C_t$ contains agent-written knowledge.
Below, we discuss how these components support complementary forms of memory.

\subsection{Memory Representation}
\label{sec:representation}
\looseness=-1
As shown in Figure~\ref{fig:overview}, MFM divides memory into three parts: episodic records of multimodal experience, a searchable index, and agent-written memory. This approach is based on the difference between episodic memory, which relates to specific experiences, and semantic and procedural memory, which concern knowledge and ways of acting~\citep{tulving1972episodic,squire2004memory}. The episodic store $E_t$ holds the observations, states, and actions, while the index $I_t$ gives cues for finding these records. The agent-written memory $C_t$ links discoveries from different episodes and keeps environmental knowledge, reusable procedures, and working plans. The following section explains each of these in turn.

\begin{figure*}[t]
  \centering
  \includegraphics[width=\linewidth]{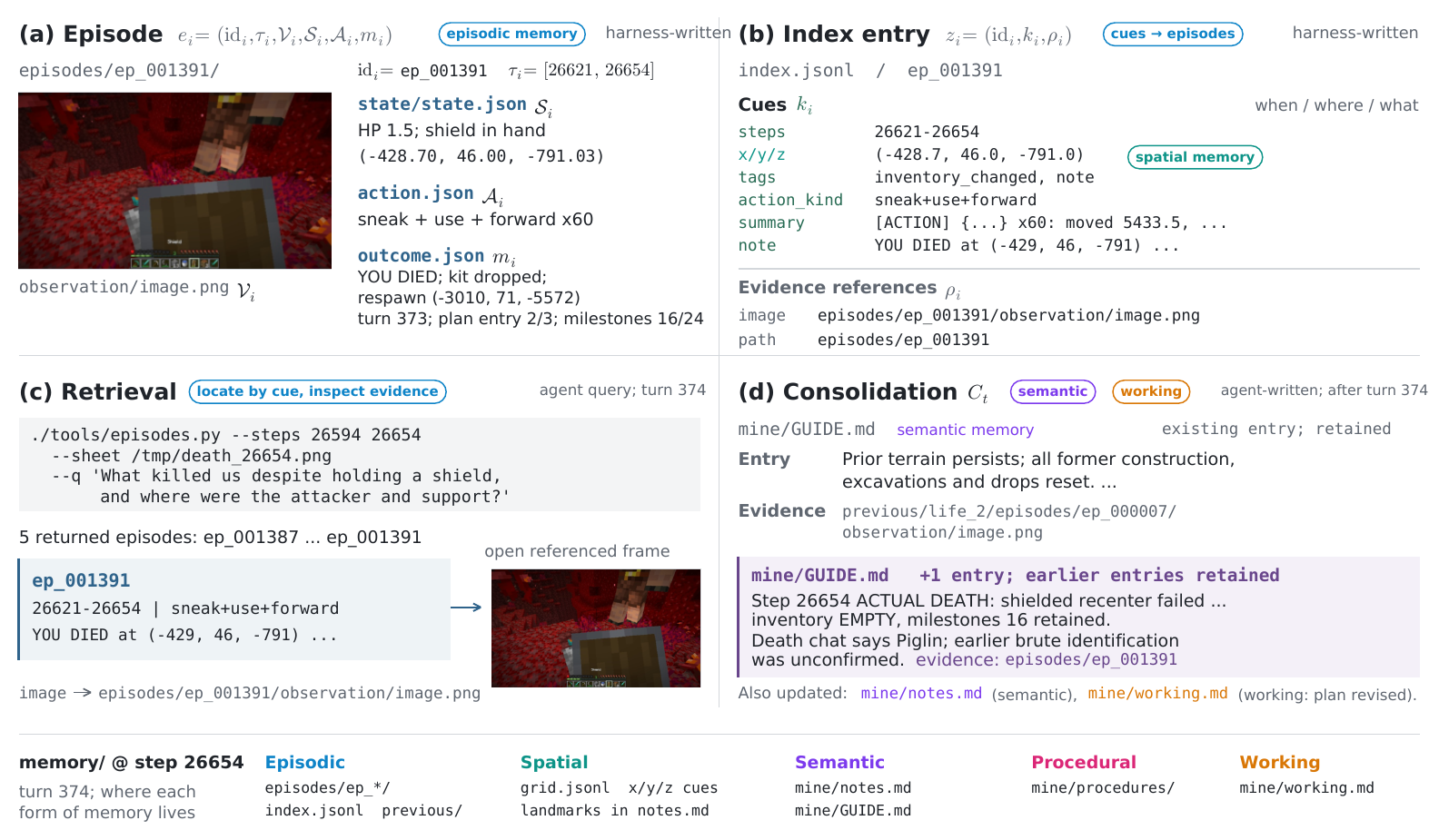}
  \vspace{-20pt}
  \caption{\small \textbf{Memory contents and operations in MFM.} A Minecraft trajectory shows how each form of memory is populated and used. \textbf{(a)} An episode: observation, state, action, and outcome stored as persistent files. \textbf{(b)} Its index entry: searchable cues, including spatial coordinates, with references to the records. \textbf{(c)} Retrieval: the agent locates episodes by step and inspects the referenced frame. \textbf{(d)} Consolidation: the evidence updates semantic (\texttt{GUIDE.md}) and working memory while earlier entries and episode references are preserved. Bottom: where each form of memory lives at this step.
  }
  \label{fig:memory-detail}
  \vspace{-15pt}
\end{figure*}

\textbf{Episodic memory.}
To revisit a past experience, the agent needs access to what it observed, what it attempted, and what happened. We retain episodes as multimodal records,
\[
e_i=\bigl(\mathrm{id}_i,\tau_i,\mathcal{V}_i,\mathcal{S}_i,\mathcal{A}_i,m_i\bigr),
\]
where $\mathrm{id}_i$ is the episode identifier, $\tau_i=[b_i,d_i]$ is the episode's timestep interval, $\mathcal{V}_i$ contains recorded visual observations, $\mathcal{S}_i$ contains associated structured states, $\mathcal{A}_i$ records actions, and $m_i$ contains metadata such as plan progress and task milestones. Episode identifiers and timestep metadata link these records to the corresponding agent interactions. Images retain perceptual details that may be absent from a textual description, while states, actions, and outcomes allow the agent to reconstruct the circumstances of an event. Preserving these records allows the agent to inspect details whose relevance becomes apparent only later. \Cref{fig:memory-detail}(a) illustrates the contents of an episode.

\textbf{Associative indexing.}
Episodic memory is useful only if the agent can retrieve relevant experiences when needed. MFM associates each episode with cues describing what happened, where it occurred, and when it took place. This use of cues to access detailed records is motivated by indexing accounts of episodic memory~\citep{teyler2007hippocampal}. We represent the index as
\[
I_t=(z_1,\ldots,z_{N_t}), \qquad
z_i=(\mathrm{id}_i,k_i,\rho_i),
\]
where $k_i$ contains searchable cues and $\rho_i$ contains references to the stored evidence. The cues include temporal and available location metadata, action type, retrieval tags, and a short action--outcome description. The harness maintains one entry per completed episode in \texttt{index.jsonl}.

The index serves as an \emph{addressing layer}: cues narrow the search to relevant episodes, while references provide access to their underlying observations. For example, the agent can locate an encounter near a cave through its recorded location, an object tag, or an attempted action, then inspect the corresponding image. This allows the agent to retrieve an experience through its context without first recalling its precise visual content. Details absent from the index remain available for inspection in the stored observations. \Cref{fig:memory-detail}(b) illustrates this separation between cues and evidence, and \Cref{alg:mfm-record} summarizes episode storage and indexing.

\textbf{Spatial memory.}
For embodied interaction, location provides a common frame for connecting experiences collected at different times. MFM supports this function through spatial information distributed across its episodic store, index, and agent-written memory. Our memory filesystem maintains an append-only grid of traversed cells in a world frame in \texttt{grid.jsonl}. Episode coordinates support retrieval by spatial proximity or vertical position, while stored egocentric views let the agent verify a retrieved location by its appearance (\Cref{fig:teaser}(a,d)). The agent further records landmark knowledge in \texttt{notes.md}, such as the entrance to a mine that has become blocked, together with coordinates and references to supporting episodes. These representations draw on complementary ideas from cognitive maps~\citep{tolman1948cognitive}, memory for events and their locations~\citep{clayton1998episodic}, and visual place recognition~\citep{cartwright1983landmark}. 


\textbf{Semantic and procedural memory.} The agent must also accumulate knowledge that is still useful after a specific episode has ended. This knowledge is stored in the agent-created collection $C_t$, which provides support for both semantic and procedural memory as well as the working memory mentioned below. Semantic memory involves recording facts and the relationships between them across different experiences. For instance, a note can link a resource to a cave, connect the cave to its coordinates, or relate an attempted action to the result of that action. These associations are given in plain text and include episode identifiers or file paths, enabling the agent to refer back to the relevant evidence. Useful discoveries and local knowledge are stored in \texttt{notes.md}, while more general environmental patterns and operating instructions are kept in \texttt{GUIDE.md}.


Procedural memory consists of methods that can be reused. In the \texttt{procedures/} directory there are textual or executable sequences of actions which the agent is able to reuse and modify during later interactions. Experience is converted into these kinds of knowledge and skill through periodic consolidation \citep{kumaran2016learning}. The agent can add to, correct or reorganize these procedures as new evidence becomes available, the underlying episodic records staying unaltered.

\textbf{Working memory.}
To continue an ongoing task across context changes, MFM maintains the agent's current plan and immediate subgoals in \texttt{working.md} (part of $C_t$). This file records the current plan, unresolved questions, and information needed to resume after a context reset. It connects the agent's accumulated experience and knowledge to the decisions required by its present objective.

\subsection{Memory Retrieval}
\label{sec:retrieval}
Retrieving an experience requires both locating a relevant episode and determining what its observations reveal about the current task. MFM separates this into two stages. The agent first searches for episodes through textual or structured cues, then inspects selected multimodal evidence. As the cues determine where to look without requiring the agent to describe every detail, the agent can ask new questions of past observations, including ones whose relevance was not apparent when the experience was recorded. \Cref{fig:memory-detail}(c) illustrates this process, and \Cref{alg:mfm-retrieve} summarizes the retrieval procedure.

\textbf{Agent-initiated search.}
Given its current information need, the agent constructs a query $q_t$ over the episode index $I_t$ and receives matching entries $R_t$. A query specifies conditions on recorded context, such as a time interval or location. For example, an agent investigating a failed action can locate episodes around the time it occurred. Search then identifies episodes that match the requested context, while the agent judges which are useful for its current decision. Each result provides the episode identifier, contextual cues, an action--outcome summary, and references to the evidence. Matches are returned in chronological order, with an optional limit that retains only the most recent matches.

\textbf{Visual inspection.}
If you find a particular episode, you gain access to information that goes beyond the description in its index. The agent then chooses the references $\mathcal{P}_t \subseteq \bigcup_{z_i\in R_t}\rho_i$ from the results and loads the relevant evidence into the context, $ D_t=\{\operatorname{Load}(p)\mid p\in\mathcal{P}_t\} $, where $\operatorname{Load}$ converts a reference into its stored frame, state, or other record. This approach of selectively loading evidence enables the agent to look at the evidence it needs without including the entire interaction history in the context. With respect to visual evidence, the agent may examine an original frame, compare a number of frames on a contact sheet, place a previous observation alongside the current view, or zoom in on a region of interest. These operations help answer different questions regarding the same experience, for example identifying an object, recognising a location, or understanding the outcome of an action. Since the observations are available independently of the textual clues, the agent is able to look at details that are missing from $k_i$ and reconsider a previous interpretation. When an inspection involves a question, the agent keeps both the question and the result so that the reason for the lookup is not lost.

\vspace{-7pt}
\subsection{Self-Evolving Consolidation}
\label{sec:consolidation}

Episodic memory preserves past experiences, while semantic and procedural memory capture knowledge and skills for future scenarios. MFM connects these forms of memory through consolidation, drawing on how complementary memory systems support learning from experience~\citep{kumaran2016learning}. During consolidation, the agent examines accumulated evidence and updates its understanding of the environment, reusable procedures, and its current plan. These updates change the agent-written memory $C_t$ while preserving the recorded history $(E_t,I_t)$. \Cref{alg:mfm-consolidate} summarizes this procedure.

\textbf{From experience to knowledge and skills.}
Consolidation allows findings from individual episodes to inform subsequent behavior. The agent can record a useful observation as local knowledge in \texttt{notes.md}, or update \texttt{GUIDE.md} when experience supports a broader regularity. For example, an encounter may establish where a resource was found, while repeated encounters may suggest conditions under which an action succeeds or fails. The agent can also qualify earlier guidance when new observations reveal limits to its applicability.


Procedural updates record effective methods of acting; the agent is able to convert a sequence of actions into a textual or executable procedure and then modify it as a result of the changes revealed by later attempts. Working-memory updates link the knowledge that has been accumulated to the current task by having the agent revise the file \texttt{working.md} to show the completed subgoals, any changes to the plan, and the questions that remain unanswered. The agent decides which of these updates to make on the basis of the evidence available and the role of each memory, and the resulting documents and procedures stay accessible during future interactions.


\textbf{Revision with preserved episodic evidence.} Knowledge gained from experience can become out of date or incorrect. By linking the knowledge that agents write to the episodic observations that support it, agents are able to go back and look at the basis of their earlier beliefs. For instance, after having noticed a sheep on a hillside the agent then decides to go back for food, enlarges the original frame, and discovers that the animal was in fact a wolf; it then corrects the note and alters its plan while still keeping the reference to the episode. If the agent had kept only the textual description it would have lost the visual details required to correct that interpretation. Therefore, by preserving the episodic evidence the agent can reevaluate its earlier conclusions as its understanding develops. Figure~\ref{fig:memory-detail}(d) shows that the agent updates its \texttt{GUIDE} to indicate that the death message names a Piglin and that the previous identification as a brute has not been confirmed.
\vspace{-10pt}
\section{Experiment Results}
\label{sec:results}
\vskip -0.2cm


We evaluate MFM to understand how persistent multimodal memory affects task performance, decision-making, and interaction efficiency. We compare MFM with baseline methods on long-horizon embodied and computer-use tasks (\Cref{sec:results}), examine the memory component contributions via ablation studies (\Cref{sec:ablation}), and illustrate qualitative retrieval behaviors (\Cref{sec:qualitative}), with a portal-building case study examining task completion and interaction efficiency, and analyze how the
agent learns and reuses skills through memory (Section~\ref{sec:skill-learning}).

\vspace{-7pt}
\subsection{Experimental Setup}
\label{sec:results-setup}
\vskip -0.2cm
We evaluate \textsc{MFM} across two diverse domains, averaging all main and ablation results over three random seeds (determining the Minecraft world or KiCad task order). We limit each configuration to three seeds because long-horizon runs with frontier models are expensive. All models are ran with \texttt{high} reasoning efforts.
\begin{itemize}[leftmargin=2em, nosep]
    \item \textbf{Minecraft}: \textbf{MCU-AgentBeats} \citep{zheng2025mcu,liu2026agentbeats}: open-world exploration and interaction in Minecraft, scored via scene-specific milestones. The details of both tasks are included in \ref{sec:task-details}. For the Mine Diamond from Scratch task, we use 12,000 max steps; for the Ender Dragon task, we use 36,000 max steps. We use a “life” to denote an attempt from spawn until death; after death, the agent’s inventory and world modifications reset.
    \item \textbf{CUABench-KiCad}. Built on the CUABench framework \citep{cua2025,cuabench_leaderboard} via low-level action spaces, this environment evaluates visual GUI agents on realistic electronic design automation workflows in KiCad.  It contains $N=25$ tasks. For MFM, VISTA, and PRO-LONG, memory can carry over across tasks.  A task may receive a score in \([0, 1]\) when the solution satisfies only a subset of the required criteria. 

\end{itemize}


\textbf{Baselines and models.} We compare MFM against direct evaluation (no persistent memory), PRO-LONG~\citep{fox2026pro}, and VISTA~\citep{vista2026} using frontier base models (GPT-5.6-Sol, Claude Opus 5, and GPT-6-Astra). For a fair comparison, all methods share identical base agent prompts, decoding settings, environments, harness, and context management. 


\begin{table*}[ht]
\centering
\vspace{-2pt}
\begin{minipage}[ht]{0.52\textwidth}
\centering
\caption{\small
\textbf{Finished milestones on long-horizon MCU tasks.}
Number of milestones completed on Mine Diamond from Scratch and
Ender Dragon. Results are mean $\pm$ std over
3 seeds; best in \textbf{bold}.
}
\label{tab:mcu}
\footnotesize
\setlength{\tabcolsep}{3pt}
\begin{adjustbox}{max width=\linewidth}
\begin{tabular}{@{}llcc@{}}
\toprule
Method & Model & Diamond & Dragon \\
& & (Max: 10) & (Max: 24) \\
\midrule
Direct
    & GPT-5.6-Sol & 4.0 \(\pm\) 1.0 & 6.0 \(\pm\) 1.0 \\
Evaluation
    & Claude Opus 5 & 6.0 \(\pm\) 1.0 & 6.0 \(\pm\) 1.0 \\
    & GPT-6-Astra & 6.3 \(\pm\) 0.6 & 6.3 \(\pm\) 0.6 \\
\midrule
PRO-LONG
    & GPT-5.6-Sol & 8.0 \(\pm\) 1.0 & 8.3 \(\pm\) 0.6 \\
\citep{fox2026pro}
    & Claude Opus 5 & 8.0 \(\pm\) 1.0 & 8.0 \(\pm\) 1.0 \\
    & GPT-6-Astra & 9.3 \(\pm\) 0.6 & 11.0 \(\pm\) 1.0 \\
\midrule
VISTA
    & GPT-5.6-Sol & 9.3 \(\pm\) 0.6 & 9.0 \(\pm\) 1.0 \\
\citep{vista2026}
    & Claude Opus 5 & 8.7 \(\pm\) 0.6 & 8.7 \(\pm\) 0.6 \\
    & GPT-6-Astra & 9.7 \(\pm\) 0.6 & 12.7 \(\pm\) 0.6 \\
\midrule
\textsc{MFM} (ours)
    & GPT-5.6-Sol & 9.7 \(\pm\) 0.6 & 10.7 \(\pm\) 0.6 \\
    & Claude Opus 5 & 9.3 \(\pm\) 0.6 & 10.3 \(\pm\) 0.6 \\
    & GPT-6-Astra & \textbf{10.0} \(\pm\) 0.0 & \textbf{16.0} \(\pm\) 0.0 \\
\bottomrule
\end{tabular}
\end{adjustbox}
\end{minipage}
\hfill
\begin{minipage}[ht]{0.45\textwidth}
\centering
\caption{\small
\textbf{Mean reward on CUABench-KiCad.}
Results are mean $\pm$ std over 3 seeds; best in \textbf{bold}.
}
\label{tab:kicad}
\footnotesize
\setlength{\tabcolsep}{3pt}
\begin{adjustbox}{max width=\linewidth}
\begin{tabular}{@{}llc@{}}
\toprule
Method & Model & Mean Reward \\
\midrule
Direct
    & GPT-5.6-Sol & 0.240 \(\pm\) 0.026 \\
Evaluation
    & Claude Opus 5 & 0.240 \(\pm\) 0.023 \\
    & GPT-6-Astra & 0.267 \(\pm\) 0.021 \\
\midrule
PRO-LONG
    & GPT-5.6-Sol & 0.306 \(\pm\) 0.014 \\
\citep{fox2026pro}
    & Claude Opus 5 & 0.300 \(\pm\) 0.016 \\
    & GPT-6-Astra & 0.375 \(\pm\) 0.012 \\
\midrule
VISTA
    & GPT-5.6-Sol & 0.400 \(\pm\) 0.011 \\
\citep{vista2026}
    & Claude Opus 5 & 0.373 \(\pm\) 0.013 \\
    & GPT-6-Astra & 0.400 \(\pm\) 0.010 \\
\midrule
\textsc{MFM}
    & GPT-5.6-Sol & 0.400 \(\pm\) 0.009 \\
(no cross-task
    & Claude Opus 5 & 0.375 \(\pm\) 0.010 \\
memory)
    & GPT-6-Astra & 0.413 \(\pm\) 0.009 \\
\midrule
\textsc{MFM} (ours)
    & GPT-5.6-Sol & 0.413 \(\pm\) 0.009 \\
    & Claude Opus 5 & 0.389 \(\pm\) 0.010 \\
    & GPT-6-Astra & \textbf{0.440} \(\pm\) 0.013 \\
\bottomrule
\end{tabular}
\end{adjustbox}
\end{minipage}

\vspace{-5pt}
\end{table*}

\subsection{Performance on Long-Horizon Tasks}
\label{sec:results}
\vskip -0.2cm

\textbf{MCU-AgentBeats}. \Cref{tab:mcu} shows completion counts. MFM + GPT-6-Astra finishes all 10 Diamond milestones (vs.\ 6.3 for direct evaluation, 9.3 for PRO-LONG, and 9.7 for VISTA). On the harder Dragon task, MFM achieves 16/24 milestones, outperforming direct evaluation (6.3), PRO-LONG (11.0), and VISTA (12.7). We observe that every baseline plateaus, making no further progress after at most 13 milestones. MFM consistently improves performance across all base models.


\textbf{CUABench-KiCad}. \Cref{tab:kicad} indicates that MFM’s benefits extend to computer-use tasks, yielding mean rewards of 0.413 (GPT-5.6-Sol), 0.389 (Opus 5), and 0.440 (GPT-6): improving over direct evaluation baselines by 0.149–0.173 and outperforming memory baselines across models. Without cross-task memory, MFM's reward drops by $0.013$--$0.027$ across models. 

\textbf{Temporal reach and interaction efficiency.} \Cref{fig:retrieval-lag} reveals that MFM accesses much earlier history than VISTA (which favors recent experience) or PRO-LONG, demonstrating the effectiveness of MFM's addressing layer. Furthermore, \cref{fig:steps-per-turn} shows that MFM and PRO-LONG execute more actions per turn than VISTA,  with fewer total model queries and reducing execution overhead.
\begin{table}[t]
    \centering
    \small
    \caption{\small \textbf{Ablation of MFM’s memory components.} Effects of removing stored frames, associative indexing, agent-written procedures or consolidation on the Diamond and Dragon tasks.
    }
    \label{tab:ablation}
    \setlength{\tabcolsep}{4pt}
    \vspace{-5pt}
    \begin{tabular}{l*{4}{cc}}
        \toprule
         & \multicolumn{2}{c}{No visual obs.}
          & \multicolumn{2}{c}{Flat}
          & \multicolumn{2}{c}{No agent procedures} 
          & \multicolumn{2}{c}{No consolidation} \\
        \cmidrule(lr){2-3}
        \cmidrule(lr){4-5}
        \cmidrule(lr){6-7}
        \cmidrule(lr){8-9}
          Model
          & Diamond & Dragon
          & Diamond & Dragon
          & Diamond & Dragon 
          & Diamond & Dragon \\
        \midrule
         GPT-5.6-Sol
          & $8.0 \pm 1.0$ & $ 8.3 \pm 0.6$
          & $8.7 \pm 0.6$ & $ 8.7 \pm 0.6$
          & $8.7 \pm 0.6$ & $ 9.0 \pm 1.0$ 
          & $8.3 \pm 0.6$ & $ 8.3 \pm 0.6$ \\
        GPT-6-Astra
          & $8.7 \pm 0.6$ & $9.7 \pm 0.6$
          & $9.3 \pm 0.6$ & $9.3 \pm 0.6$
          & $8.7 \pm 0.6$ & $10.7 \pm 1.2$ 
          & $8.7 \pm 0.6$ & $9.3 \pm 0.6$ \\
        \bottomrule
    \end{tabular}
    \vspace{-10pt}
\end{table}

\vspace{-4pt}
\subsection{Ablation Studies}
\label{sec:ablation}

We evaluate the contributions of MFM's memory components through four ablations that separately remove access to historical visual observations, associative indexing, agent-written procedures, and consolidation. \Cref{tab:ablation} reports the ablation results. \textbf{Historical visual observations} appear essential for resolving perceptual ambiguities and spatial orientation (\Cref{sec:qualitative}). Disabling stored frames drops completion significantly across models: for GPT-5.6-Sol, scores fall to $8.0 \pm 1.0$ (Diamond) and $8.3 \pm 0.6$ (Dragon), while GPT-6-Astra drops to $8.7 \pm 0.6$ (Diamond) and $9.7 \pm 0.6$ (Dragon).  \textbf{Associative indexing} suggests that a dedicated addressing layer helps; Replacing it with chronological text matching over a flat log (Flat) reduces performance to $8.7 \pm 0.6$ on both tasks for GPT-5.6-Sol, and to $9.3 \pm 0.6$ on both tasks for GPT-6-Astra. Additionally, \textbf{agent-written procedures} seem to help compound past successes into reusable routines (\Cref{sec:skill-learning}), as disabling them decreases performance for GPT-5.6-Sol (Diamond: $8.7 \pm 0.6$; Dragon: $9.0 \pm 1.0$) and more sharply for GPT-6-Astra (Diamond: $8.7 \pm 0.6$; Dragon: $10.7 \pm 1.2$). Finally,  the removal of \textbf{consolidation} lowers performance to $8.3 \pm 0.6$ on both tasks for GPT-5.6-Sol, and to $8.7 \pm 0.6$ (Diamond) and $9.3 \pm 0.6$ (Dragon) for GPT-6-Astra. The consistent drops across all four ablations is the main evidence that each component contributes.

\vspace{-4pt}
\subsection{How Memory Retrieval Helps in Long-Horizon Tasks}
\label{sec:qualitative}
\vskip -0.1cm
We examine qualitative examples to understand how MFM supports decisions during long-horizon interaction. These examples illustrate how agents inspect visual evidence to revise earlier interpretations, recover spatial context from previous observations, and reuse evidence across tasks.

\textbf{MCU-AgentBeats}. \emph{Visual inspection enables agents to correct uncertain observations.} In \Cref{fig:crop-wolf}(a,b), magnifying ambiguous frames reveals a wolf instead of a sheep, preventing a costly attack; in (c,d), pink sheep observations guide resource gathering.
\emph{Retrieving earlier observations also helps the agent recover spatial context and connect past experience to the current environment.} Agents utilize past frames to locate hole rims (\cref{fig:03}), identify mine entrances (\cref{fig:04}), and correct unfavorable viewing angles (\cref{fig:05}).

\begin{figure*}[ht]
  \centering
  \begin{subfigure}[t]{0.32\textwidth}
    \centering
    \includegraphics[width=\linewidth]{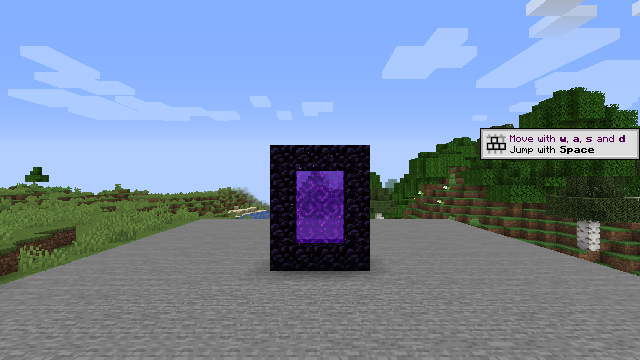} 
    \caption{\small MFM completes the portal.}
    \label{fig:}
  \end{subfigure}
  \begin{subfigure}[t]{0.32\textwidth}
    \centering
    \includegraphics[width=\linewidth]{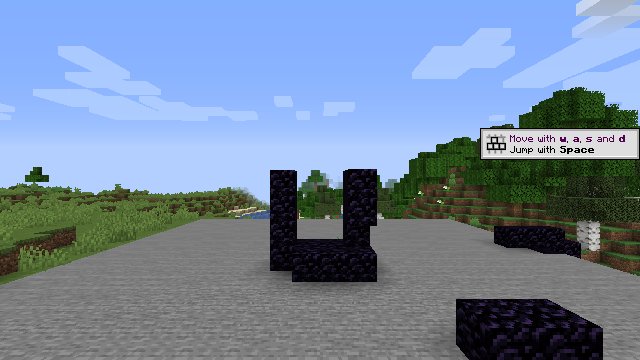} 
    \caption{\small VISTA}
    \label{fig:}
   \end{subfigure}
   \begin{subfigure}[t]{0.32\textwidth}
    \centering
    \includegraphics[width=\linewidth]{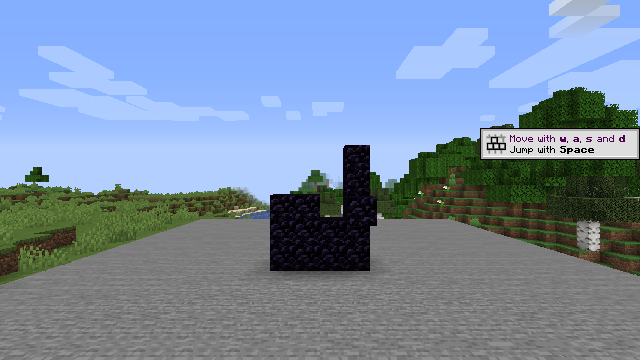} 
    \caption{\small PRO-LONG}
    \label{fig:}
   \end{subfigure}
  \caption{\small \textbf{Qualitative behavior on the Nether portal task.} MFM successfully completes the Nether portal, whereas VISTA and PRO-LONG fail to finish the structure.
  }
  \label{fig:portal-case}
  \vspace{-20pt}
\end{figure*}
\begin{figure*}[ht]
    \centering
    \begin{minipage}[b]{0.42\textwidth}
        \centering
        \includegraphics[height=4.2cm,width=\linewidth,keepaspectratio]
            {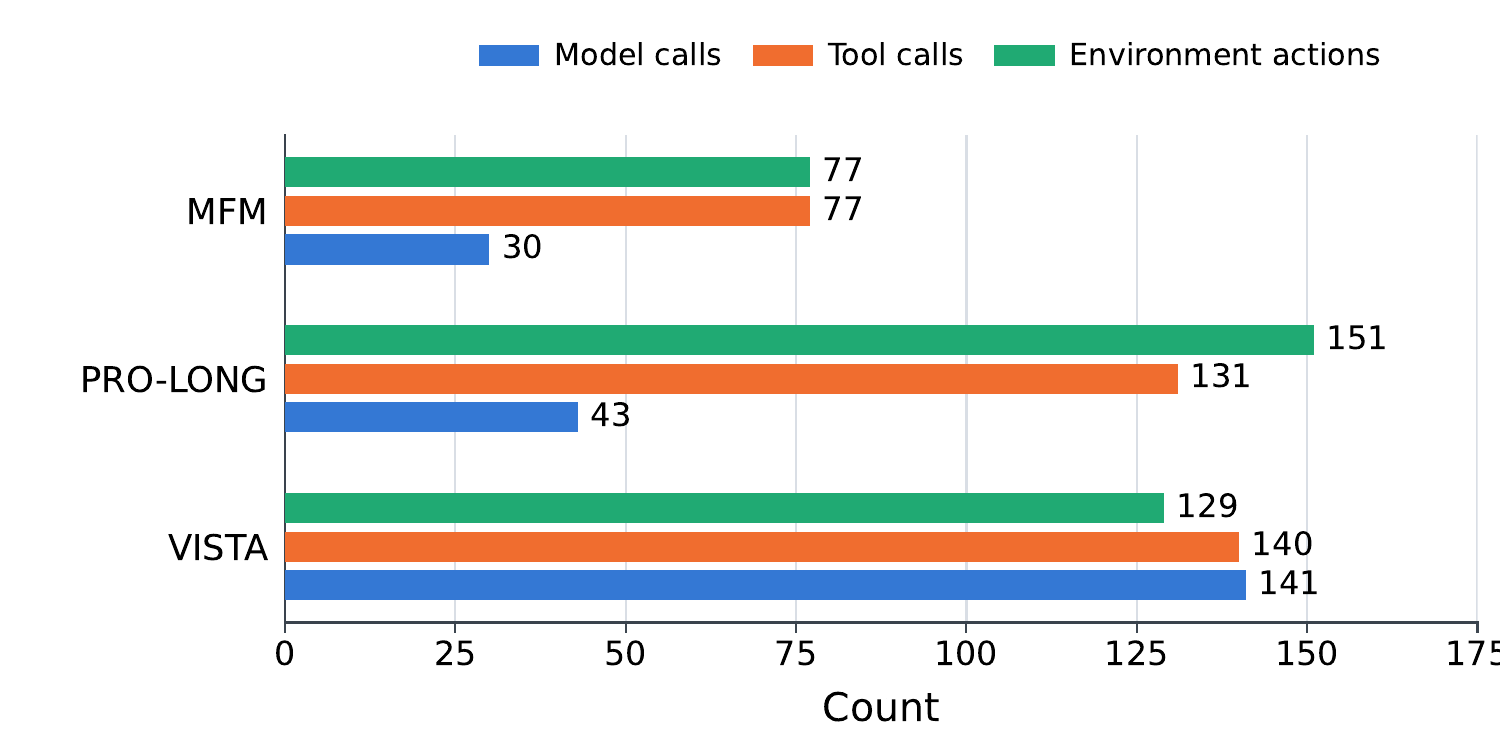}
    \end{minipage}\hfill
    \begin{minipage}[b]{0.56\textwidth}
        \centering
        \includegraphics[height=4.2cm,width=\linewidth,keepaspectratio]
            {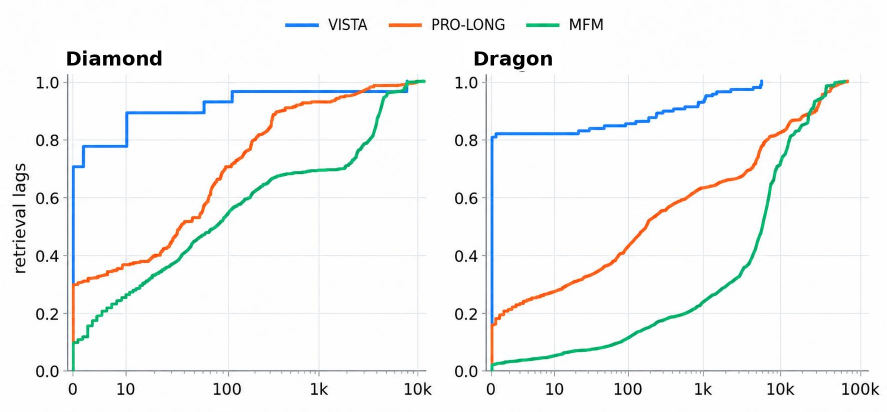}
    \end{minipage}
    \vskip -0.2cm
    \begin{minipage}[t]{0.42\textwidth}
        \captionof{figure}{\small
            \textbf{Interaction efficiency on the Nether portal task.}
            Comparison of model calls, tool calls, and executed environment actions.}
        \label{fig:portal-case-number}
    \end{minipage}\hfill
    \begin{minipage}[t]{0.56\textwidth}
        \captionof{figure}{\small
            \textbf{Temporal reach of memory retrieval.}
            Retrieval-lag (environment steps between recording an experience and retrieving it) on Diamond and Dragon. VISTA favors recent experience; MFM reaches further back, especially on Dragon.}
        \label{fig:retrieval-lag}
    \end{minipage}
    \vskip -0.6cm
\end{figure*}

We further examine a controlled construction task requiring the agent to align visual observations across multiple actions with a textual specification (full prompt in \cref{box:nether}). Starting on a cleared stone platform with sufficient resources, the agent must build and activate a Nether portal. This probes multimodal capability: the agent must translate textual geometric constraints into a 3D structure, track construction progress from first-person views, and adjust accordingly. Only MFM successfully built the portal under a 6,000-step budget (\cref{fig:portal-case}), requiring lower overhead (\cref{fig:portal-case-number}): 30 model calls, 77 tool calls, and 77 environment actions (vs.\ 141/140/129 for VISTA and 43/131/151 for PRO-LONG).

\textbf{CUABench-KiCad.} \Cref{fig:kicad-tasks-different-perf} shows 3 tasks where MFM outperforms VISTA (GPT-6-Astra). In Task 55f2eefb (switching an LED circuit's supply from 9\,V to 5\,V), both methods log an ``OK'' from Change
Symbol, but only MFM verifies the resulting netlist: its final state reads +5V
while VISTA's still reads +9V, underscoring the need to check task state through visual memory. On a buck-regulator task (c4baee36), both methods
produce identical components and wiring but different supply labels; on the
two-op-amp task (b7148878), MFM leaves 1 input pin unconnected versus 3 for
VISTA. \Cref{fig:kicad-cross-task-reuse} shows episodic retrieval across tasks: an
export sequence from one task is reused on another.


\section{Self-Improvement with MFM}

\subsection{Skill Learning through Memory}
\label{sec:skill-learning}


We look into whether long-term memory allows agents to enhance their future behaviour by turning experience from interactions into reusable procedures. In this context, self-improvement means adapting as a result of the memory accumulated in MFM. MFM includes the facilities needed to store, modify, and carry out procedures written by the agent, and it is the agent itself that decides which behaviours to keep and when to reuse them. We do not specify the resulting skill library and do not require a fixed number of skills for each run. The analysis presented in this section includes observations of skill creation and reuse over different trajectories, an ablation study of the agent-written procedures, and a controlled study of the number of steps required to solve new tasks with or without reusing the agent-written procedures.


\begin{figure}[t]
  \centering
  \includegraphics[width=\linewidth]{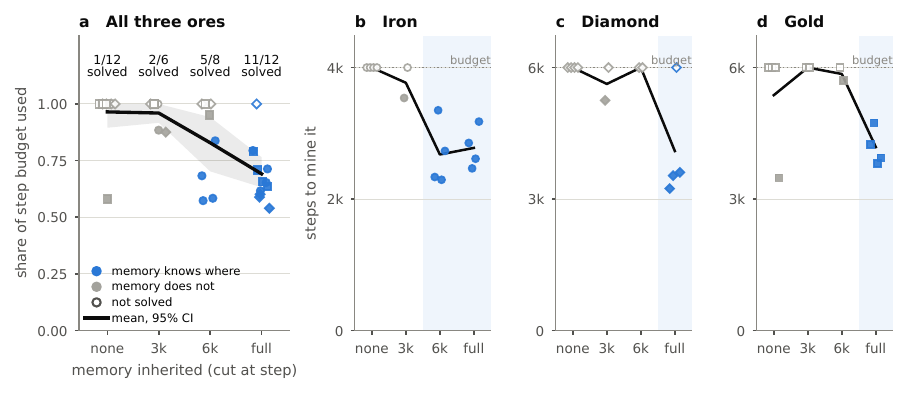}
  \vspace{-20pt}
  \caption{\textbf{More memory of the world makes new tasks faster.} We take the memory of an earlier agent that explored the world and give fresh agents progressively larger portions of it: none, an early cut, a middle cut, or the full memory. Each agent is then asked to find and mine an ore (iron, diamond or gold). (a) As more memory is inherited, tasks take fewer steps and are solved more often. The gain comes from what the memory contains. Agents whose memory already records where the needed ore lies (blue) almost always succeed and finish well within budget. Agents whose memory predates that discovery (grey) do no better than agents with no memory. (b–d) The pattern holds for each task. Informed agents dig straight to the remembered location instead of searching blindly.
  } 
  \label{fig:memory-transfer}
  \vspace{-20pt}
\end{figure}

\textbf{From experience to reusable behaviour.} The trajectories show how agents convert their experience of interaction into permanent action routines. In a run of GPT-6-Astra Dragon, after manually cutting a descending staircase (steps 5,068 to 5,249), the agent created the procedure \texttt{cut\_step}, which is parameterised by heading and cuts one step while advancing (\ref{skill:cut-step}); it then invoked this procedure 60 times, once for each of the four cardinal headings. A second procedure, \texttt{quarry\_north\_pair}, extracts the upper and lower block from the wall in front and gathers the dropped items (\ref{skill:quarry-pair}); the 27th time it was called, at step 17,462, it broke the first diamond ore block. Reuse also carries over from one life to another. The procedure \texttt{bridge\_north}, which places one floor block while sneaking and takes the target cell as input (\ref{skill:bridge-north}), was written in the previous life after the agent had manually placed two blocks over the lava and was then called three times in that same life. In this life the agent called it nine times between steps 19,261 and 20,103, using varying inputs. Each invocation placed one cobblestone block and left the agent's health unchanged, and the bridge ended up reaching the lava pool where mining of obsidian took place at step 20,736. The agent also disabled one inherited procedure, \texttt{level\_step}, having traced the previous death in the lava back to this procedure's release of the \texttt{sneak} key.


\begin{table}[ht]
    \centering
    \small
    \vspace{-5pt}
    \caption{\small \looseness=-1 \textbf{Skills learned from one task speed up new tasks.} Mean environment
    steps (over 5 seeds each) to solve three new tasks without (w.o.) and with (w.) the scripts an earlier life
    wrote installed. Every life solved its task.}
    \label{tab:skills-faster}
    \setlength{\tabcolsep}{4pt}
    \begin{tabular}{*{3}{ccc}}
        \toprule
          \multicolumn{3}{c}{Iron shield}
          & \multicolumn{3}{c}{Climb out}
          & \multicolumn{3}{c}{Cook meat} \\
        \cmidrule(lr){1-3}
        \cmidrule(lr){4-6}
        \cmidrule(lr){7-9}
          w.o. & w. & $\Delta$
          & w.o. & w. & $\Delta$
          & w.o. & w. & $\Delta$ \\
        \midrule
          2422 & 1705 & \textcolor{green!50!black}{ $\downarrow 30\%$ }
          & 2010 & 1867 & \textcolor{green!50!black}{ $\downarrow 7\%$ }
          & 640  & 476  & \textcolor{green!50!black}{ $\downarrow 26\%$ } \\
        \bottomrule
    \end{tabular}
    \vspace{-5pt}
\end{table}

\textbf{Contribution to task performance.} The results from the procedure ablation show that the procedures written by the agent do in fact contribute to task performance in MFM. For instance, when these procedures are removed for GPT-5.6-Sol, the mean milestone completion drops from 9.7 to 8.7 on Diamond and from 10.7 to 9.0 on Dragon. Our analysis points to two mechanisms that account for these improvements. Firstly, the skills allow agents to carry out the tasks more quickly; as presented in Table~\ref{tab:skills-faster}, reusing skills acquired during an earlier task decreased the average number of environment steps required to complete three new tasks: Iron shield, Climb out, and Cook meat by 30\%, 7\%, and 26\% respectively. Secondly, skills can enable agents to reach milestones that would otherwise be impossible for them to achieve, as shown in the case of the blaze-fighting skill. If there are no agent-written skills that take visual inputs (for example, \ref{skill:blaze-fight}), then none of the methods or models evaluated is able to go beyond milestone 16 to accomplish milestones such as killing a blaze or an enderman. These findings indicate that procedures enhance both the efficiency of task completion and the variety of milestones that can be achieved, even though they do not mean that each stored procedure is beneficial or that having a larger library of skills will necessarily lead to better performance.

\subsection{Transferring world knowledge to new tasks}

We next examine whether accumulated knowledge of an environment helps agents solve new objectives. We initialize fresh agents with memory from an earlier exploration trajectory, using four conditions: no inherited memory, memory collected through step 3,000, through step 6,000, or the full trajectory. Each agent is then tasked with finding and mining iron, diamond, or gold in the same world. \Cref{fig:memory-transfer} reports success and environment steps under each condition. Across the evaluated trials, success increases from 1/12 without inherited memory to 11/12 with full memory. The results suggest that the benefit depends on the information retained: agents whose inherited memory records the target ore’s location generally succeed within budget, whereas agents receiving memory from before its discovery often exhaust their budget. 
\section{Conclusion}

We introduced MFM, a multimodal filesystem memory that organizes episodic, spatial, semantic, procedural, and working memory as folders and files in one shared filesystem. Episodic records preserve raw observations, states, actions, and outcomes; a searchable index of contextual cues addresses them; and agent-written knowledge, procedures, and plans cite the episodes that support them. An agent can thus locate past experience by when, where, or what happened, inspect the original observation for details text omits, and revise what it has learned without discarding the evidence. Across Minecraft and KiCad with three base models, MFM outperforms memory-free and memory-augmented baselines, each component contributes, and memory transfers across tasks. We hope MFM can serve as a foundation for memory in long-horizon multimodal and embodied agents.

\textbf{Limitations and Future Work}. MFM improves long-horizon performance, but exploration remains hard: even GPT-6-Astra at maximum reasoning effort (from a concurrent experiment posted online) did not complete Ender Dragon. Finding better ways of coupling memory
to exploration and route planning could help agents find landmarks and avoid repeated traversal. Our evaluation is also limited to single agents in Minecraft and KiCad; shared memory across agents could combine experience from different regions or solution attempts, and broader environments would test how far the design transfers.

\clearpage

\bibliography{iclr2027_conference}
\bibliographystyle{iclr2027_conference}

\newpage
\appendix

\etocdepthtag.toc{appendix}
\etocsettocstyle{\section*{Appendix - Table of Contents}}{}
\tableofcontents
\vspace{1em}

%

\section{Additional Examples}
\label{app:examples}

\captionsetup[figure]{font=small}
\captionsetup[subfigure]{font=small}

This appendix adds results and qualitative examples to \Cref{sec:qualitative}.
\Cref{app:ex-minecraft-progress} covers milestone progression and interaction on Minecraft.
\Cref{app:ex-minecraft-visual} shows further cases of visual memory inspection.
\Cref{app:ex-kicad} gives qualitative examples on CUABench-KiCad.

\subsection{Milestone Progression and Interaction in Minecraft}
\label{app:ex-minecraft-progress}

\begin{figure}[t]
  \centering
  \includegraphics[width=\linewidth]{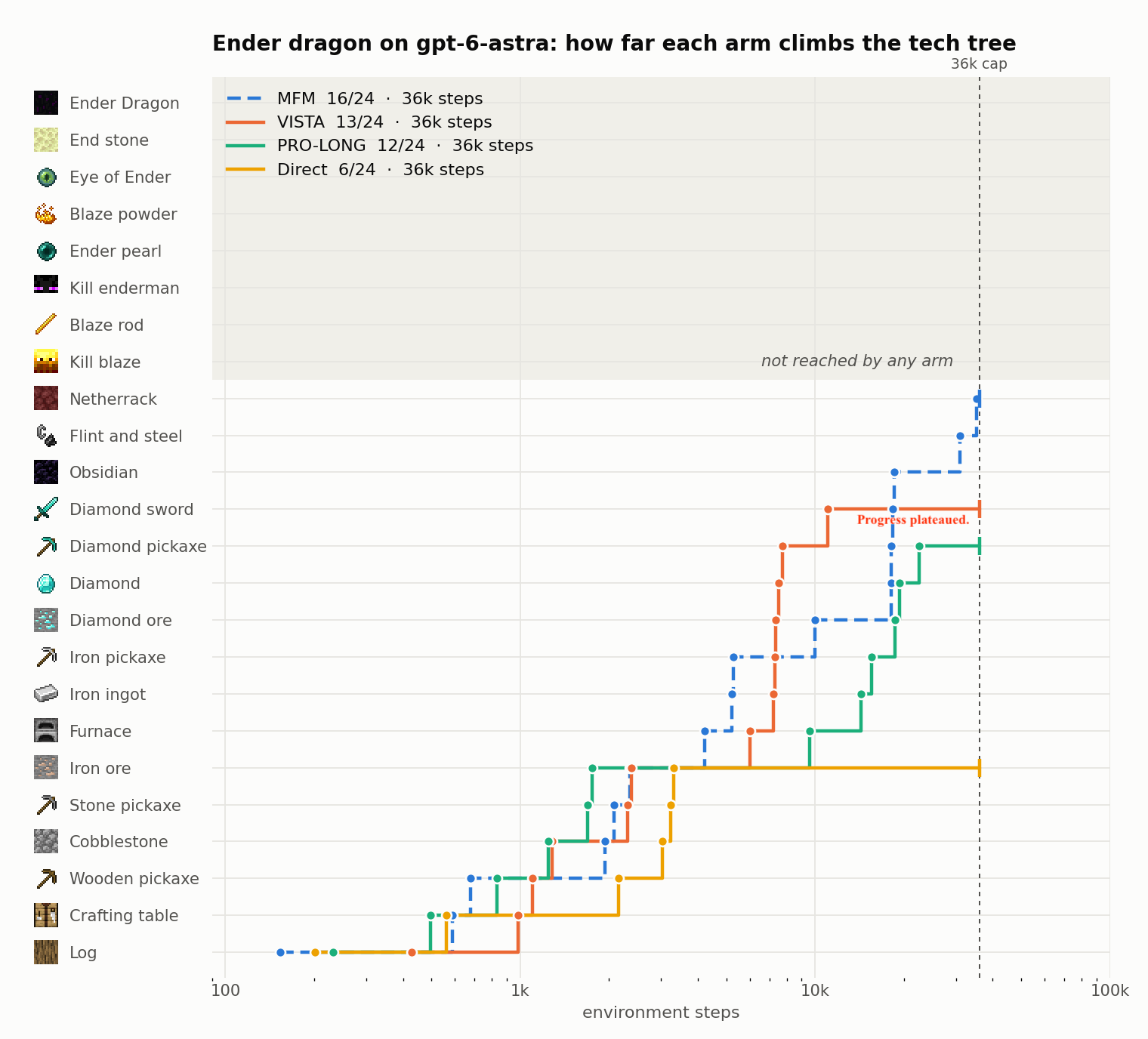}
  \caption{\textbf{Milestone progression on Kill the Ender Dragon with GPT-6-Astra.}
  Each curve shows one run's progression through the Minecraft technology tree against environment steps (log scale).
  Within the 36,000-step budget (dashed line), MFM reaches 16 of 24 milestones, compared with 13 for VISTA, 12 for PRO-LONG, and 6 for Direct Evaluation.
  The shaded region marks milestones that no method reached.
  One run is shown per method; 
  \Cref{tab:mcu} reports seed-averaged results.} 
  \label{fig:tech-tree}
\end{figure}

\Cref{fig:tech-tree} tracks milestone progression on Kill the Ender Dragon with GPT-6-Astra, using one run per method.
VISTA gets diamond equipment earlier than MFM, but its progress then stalls.
MFM continues through obsidian and flint and steel and reaches netherrack within the same 36,000-step budget.
\Cref{tab:mcu} reports seed-averaged results for all methods and base models. 

\begin{figure}[t]
  \centering
  \includegraphics[width=\linewidth]{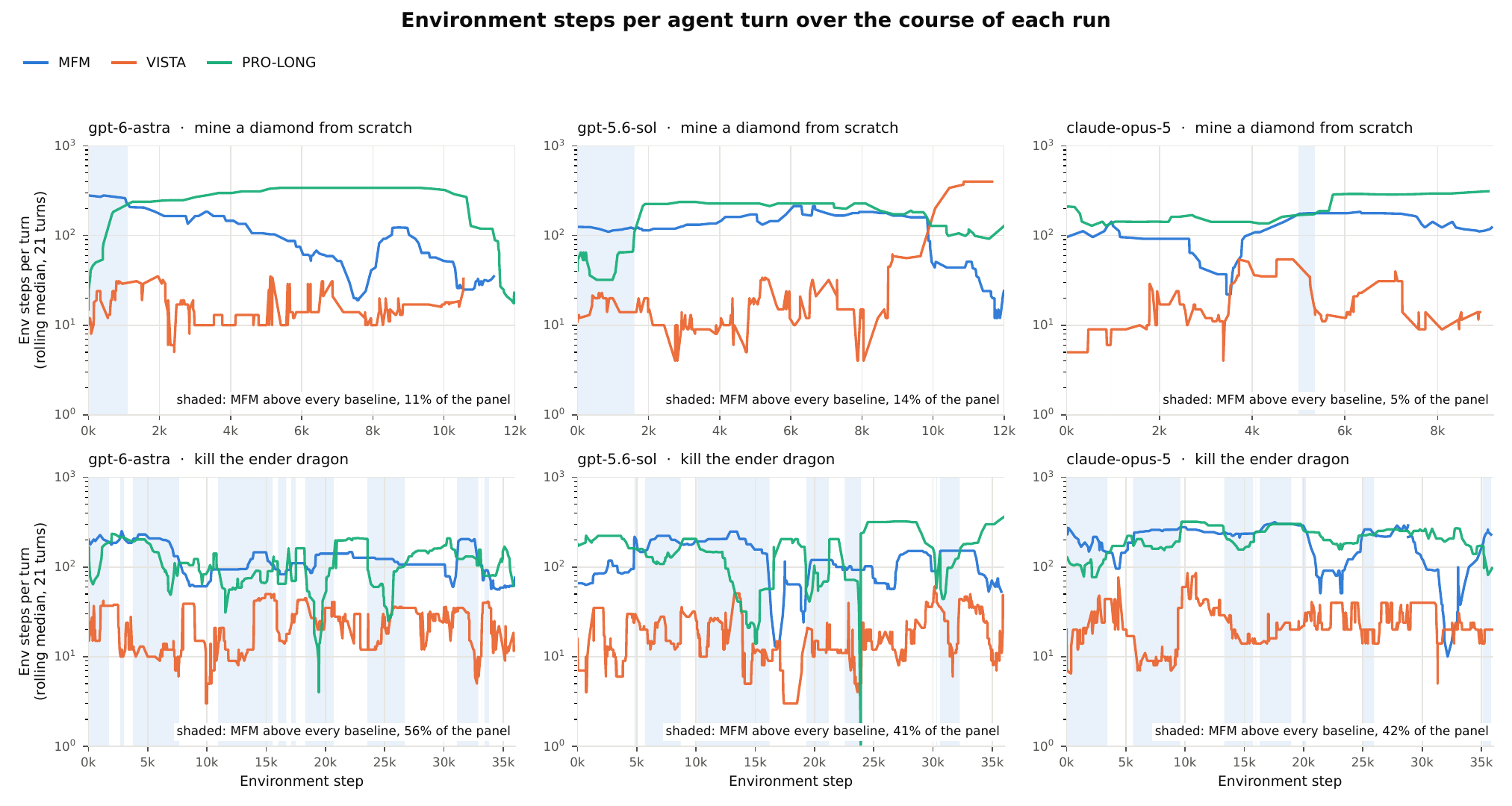}
  \caption{\textbf{Environment interaction per agent turn.}
  Rolling median of environment steps executed per agent turn over each run.
  Columns are the three base models; rows are Mine Diamond from Scratch (top) and Kill the Ender Dragon (bottom).
  Higher values mean more environment interaction between successive model calls; the vertical axis is logarithmic.
  Shaded intervals mark where MFM exceeds both VISTA and PRO-LONG, and the annotation in each panel gives the shaded fraction.}
  \label{fig:steps-per-turn}
\end{figure}

\Cref{fig:steps-per-turn} shows how many environment steps each method executes between successive model calls.
MFM and PRO-LONG generally execute more steps per turn than VISTA.
In other words, they issue longer action plans and call the model less often.
How MFM compares with PRO-LONG depends on the task.
On Ender Dragon, MFM executes more steps per turn than both baselines for 41--56\% of each run.
On Mine Diamond, it does so for only 5--14\% of each run, and PRO-LONG usually executes the most steps per turn.

\subsection{Visual Memory Inspection in Minecraft}
\label{app:ex-minecraft-visual}

\begin{figure}[t]
  \centering
  \begin{subfigure}[t]{0.56\textwidth}
    \centering
    \includegraphics[height=3.6cm,width=\linewidth,keepaspectratio]{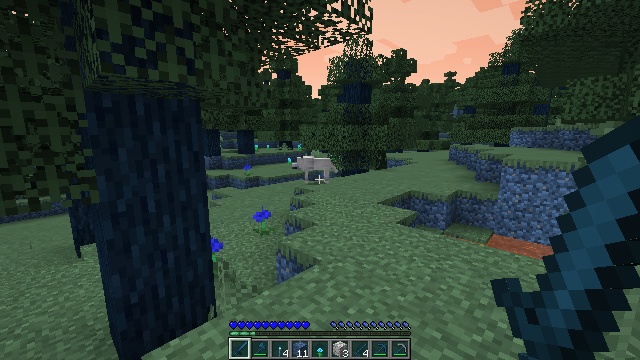}
    \caption{Stored frame: a pale animal on the hillside.}
    \label{fig:crop-wolf-a}
  \end{subfigure}\hfill
  \begin{subfigure}[t]{0.40\textwidth}
    \centering
    \includegraphics[height=3.6cm,width=\linewidth,keepaspectratio]{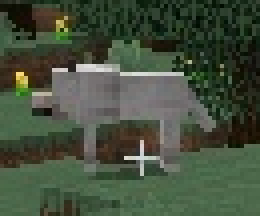}
    \caption{Magnified crop: a wolf.}
    \label{fig:crop-wolf-b}
  \end{subfigure}

  \vspace{0.5em}
  \begin{subfigure}[t]{0.56\textwidth}
    \centering
    \includegraphics[height=3.6cm,width=\linewidth,keepaspectratio]{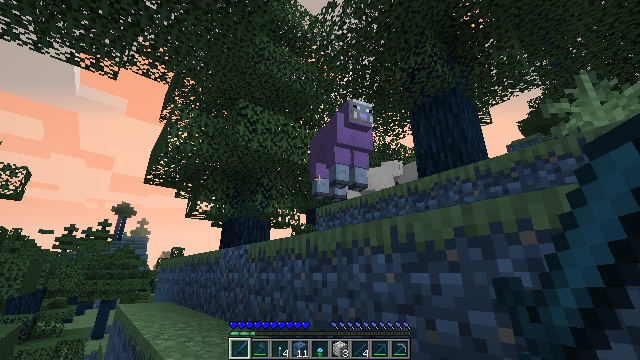}
    \caption{Stored frame: a possible pink sheep.}
    \label{fig:crop-sheep-a}
  \end{subfigure}\hfill
  \begin{subfigure}[t]{0.40\textwidth}
    \centering
    \includegraphics[height=3.6cm,width=\linewidth,keepaspectratio]{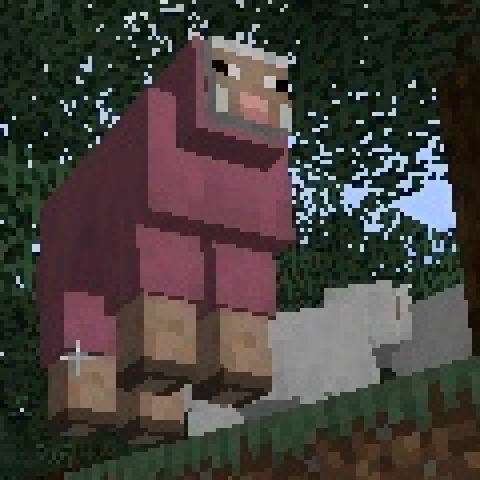}
    \caption{Magnified crop: a pink sheep.}
    \label{fig:crop-sheep-b}
  \end{subfigure}
  \caption{\textbf{Correcting and confirming perception with visual memory.}
  (a--b)~At step 4,588, the agent records a pale animal on the hillside.
  Magnifying the stored frame shows a wolf rather than a sheep, so the agent does not attack and looks for food elsewhere.
  (c--d)~At step 4,917, the agent is unsure whether an animal on the slope is a pink sheep.
  Magnifying the stored frame confirms a sheep rather than a pig, and the agent hunts it for food.}
  \label{fig:crop-wolf}
\end{figure}

\begin{figure}[t]
  \centering
  \begin{subfigure}[t]{0.34\textwidth}
    \centering
    \includegraphics[width=\linewidth]{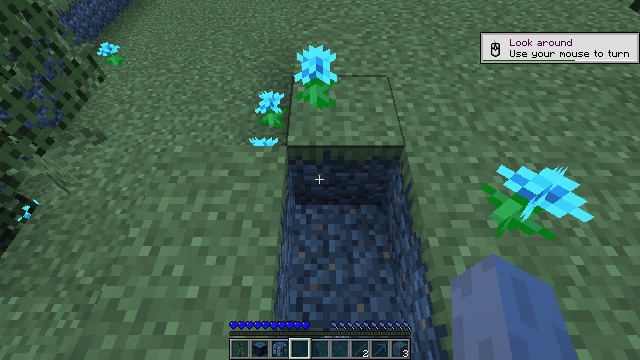}
    \caption{Stored frame of the hole the agent dug.}
    \label{fig:03-a}
  \end{subfigure}\hfill
  \begin{subfigure}[t]{0.64\textwidth}
    \centering
    \includegraphics[width=\linewidth]{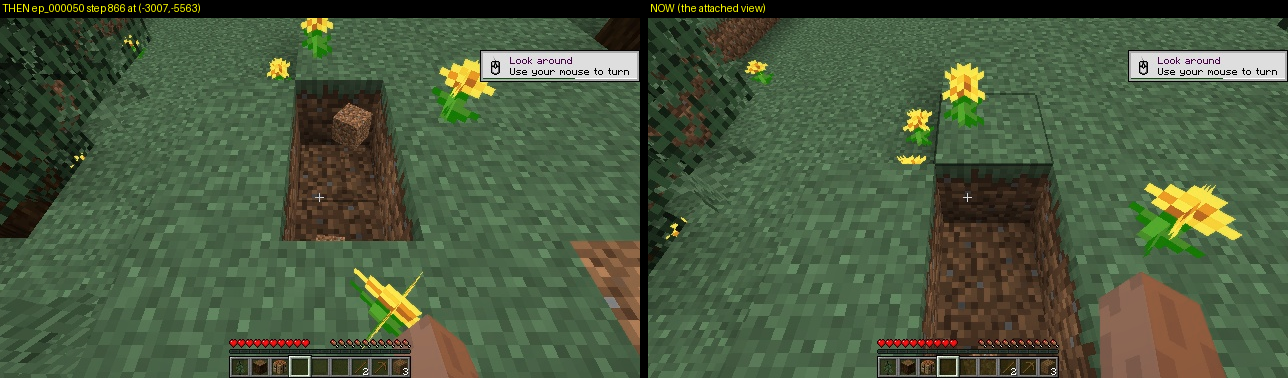}
    \caption{Retrieved frame (left) beside the current view (right).}
    \label{fig:03-then-now}
  \end{subfigure}
  \caption{\textbf{Memory-guided spatial relocation.}
  The retrieved and current views line up closely, which lets the agent identify the hole's rim and recover its exact location.}
  \label{fig:03}
\end{figure}

\begin{figure}[t]
  \centering
  \begin{subfigure}[t]{0.34\textwidth}
    \centering
    \includegraphics[width=\linewidth]{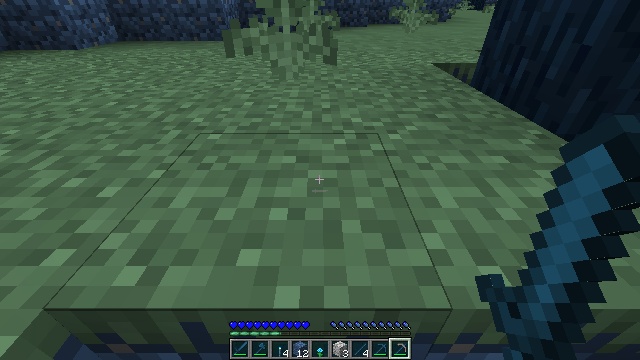}
    \caption{Life 1: stored frame of a shallow iron-mine entrance.}
    \label{fig:04_frame_640x360_step6138}
  \end{subfigure}\hfill
  \begin{subfigure}[t]{0.64\textwidth}
    \centering
    \includegraphics[width=\linewidth]{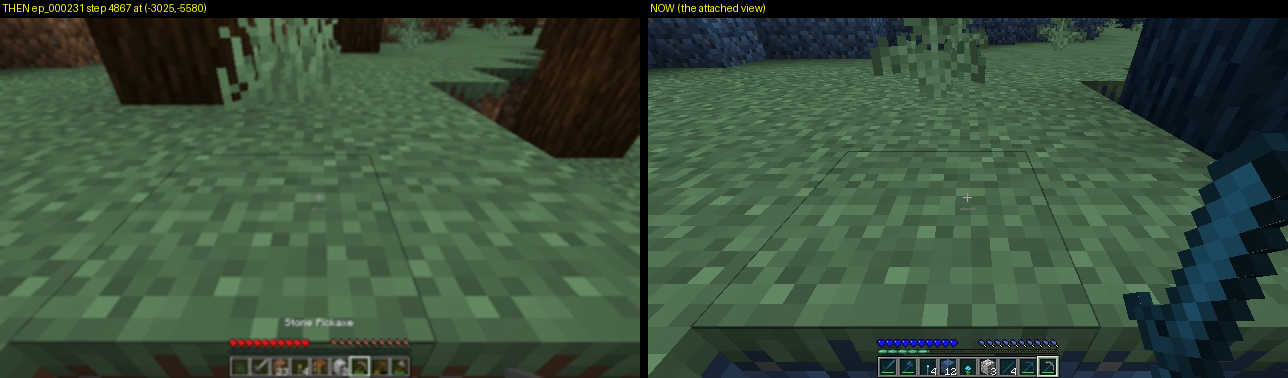}
    \caption{Life 2: retrieved frame (left) beside the current view (right).}
    \label{fig:04_then_now_sheet}
  \end{subfigure}
  \caption{\textbf{Cross-life visual memory retrieval.}
  After dying, the agent retrieves a frame recorded in an earlier life and compares it with its current view.
  It recognizes the grass bank ahead as the entrance to that life's iron mine, digs in, and mines iron ore at step 6,565.}
  \label{fig:04}
\end{figure}

\begin{figure}[t]
  \centering
  \begin{subfigure}[t]{0.49\textwidth}
    \centering
    \includegraphics[width=\linewidth]{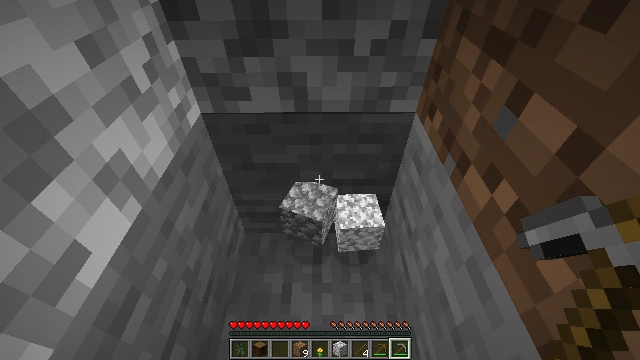}
    \caption{Retrieved frame: the earlier cut was pitched upward.}
    \label{fig:05_viewed_frame_ep_000126_step1777}
  \end{subfigure}\hfill
  \begin{subfigure}[t]{0.49\textwidth}
    \centering
    \includegraphics[width=\linewidth]{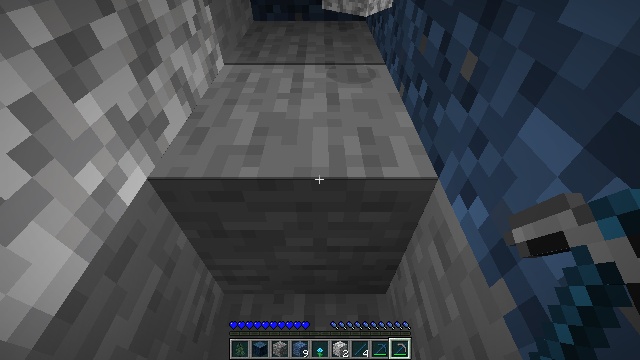}
    \caption{Re-aimed view facing the mining face.}
    \label{fig:05_frame_640x360_step1855}
  \end{subfigure}
  \caption{\textbf{Using visual memory to recover mining orientation.}
  The agent inspects the frame of an earlier cut that failed to open a level passage.
  It sees that the view was pitched upward rather than at the mining face.
  It then re-aims with \texttt{look\_abs} and resumes mining at the correct orientation.}
  \label{fig:05}
\end{figure}

\Cref{fig:crop-wolf,fig:03,fig:04,fig:05} expand on the Minecraft examples in \Cref{sec:qualitative}.
In each case, the agent finds a stored frame through its index entry.
It then inspects the frame to answer a question that the textual record alone cannot:
what kind of animal it saw (\Cref{fig:crop-wolf}),
where a hole it dug earlier is located (\Cref{fig:03}),
whether a place seen in an earlier life matches the current view (\Cref{fig:04}),
and why an earlier mining cut failed (\Cref{fig:05}).

\subsection{Qualitative Examples on CUABench-KiCad}
\label{app:ex-kicad}

\begin{figure}[t]
  \centering
  \begin{subfigure}[t]{0.32\textwidth}
    \centering
    \includegraphics[width=\linewidth]{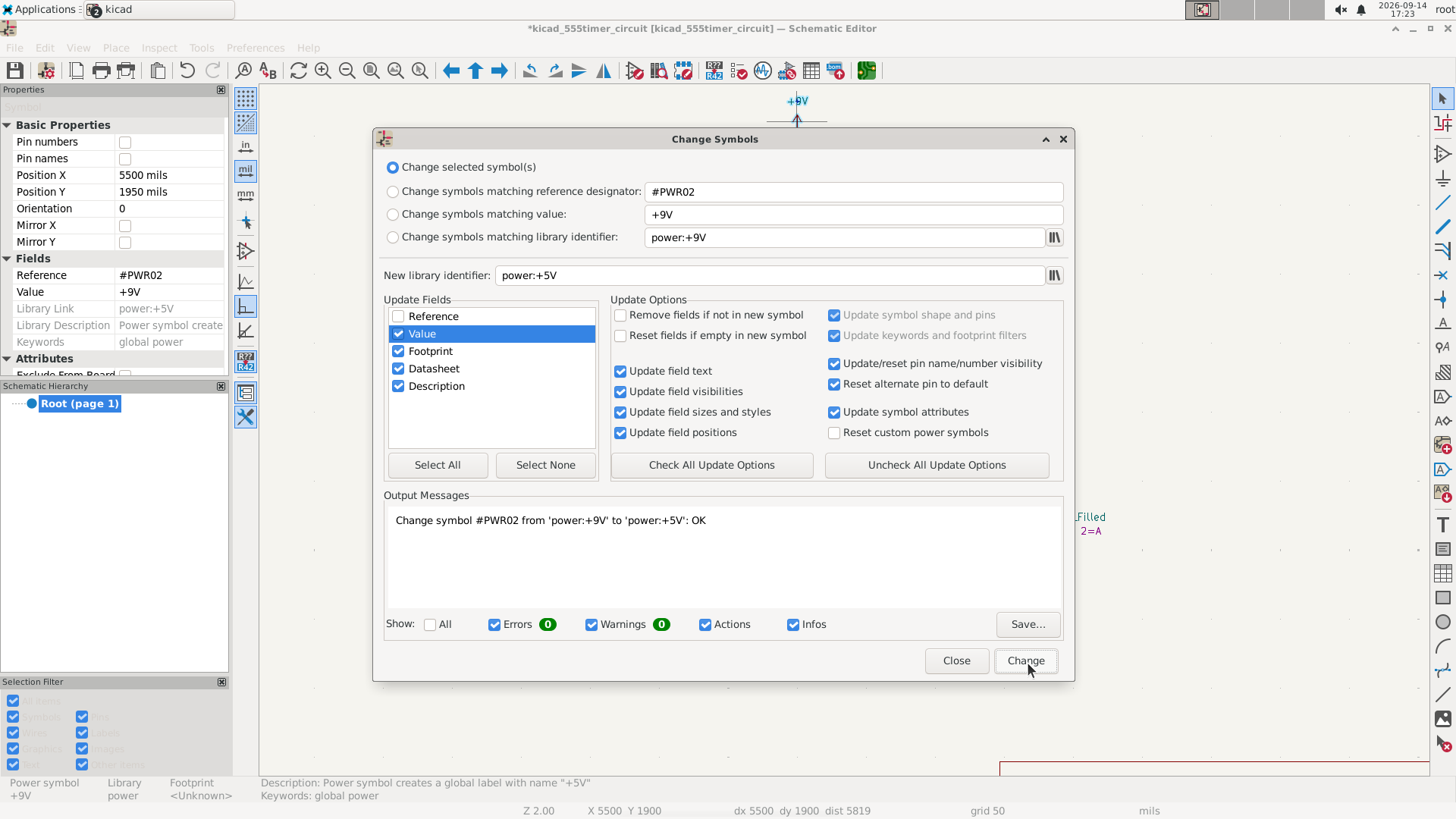}
    \caption{MFM, 55f2eefb: supply reads +5\,V.}
    \label{fig:s009_change_symbol_ok_but_agent_notices_value}
  \end{subfigure}\hfill
  \begin{subfigure}[t]{0.32\textwidth}
    \centering
    \includegraphics[width=\linewidth]{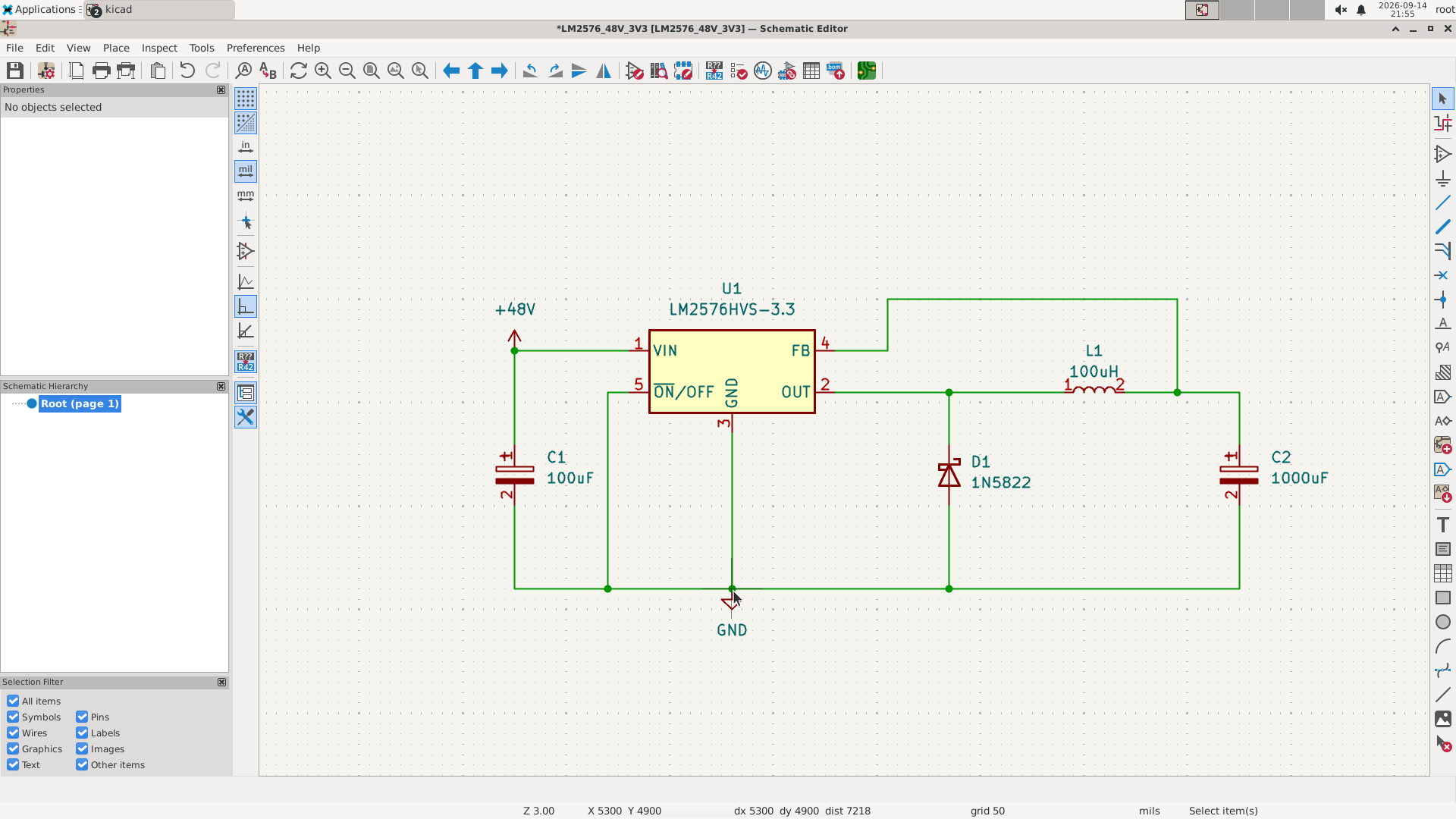}
    \caption{MFM, c4baee36: +48\,V and GND supply symbols.}
    \label{fig:s045_finished_schematic_with_48v_and_gnd_symb}
  \end{subfigure}\hfill
  \begin{subfigure}[t]{0.32\textwidth}
    \centering
    \includegraphics[width=\linewidth]{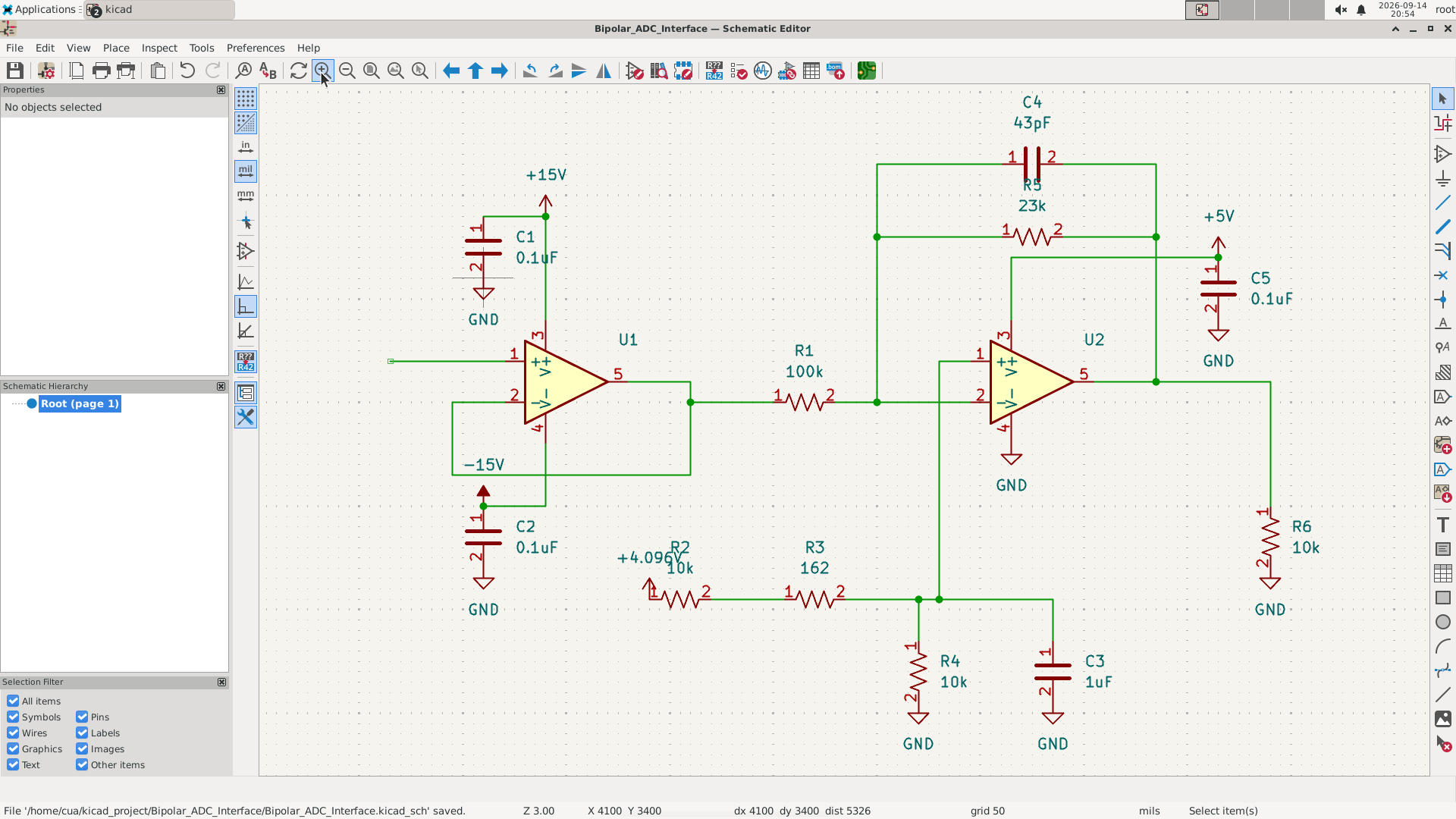}
    \caption{MFM, b7148878: one unconnected pin.}
    \label{fig:s094_full_schematic}
  \end{subfigure}

  \vspace{0.5em}
  \begin{subfigure}[t]{0.32\textwidth}
    \centering
    \includegraphics[width=\linewidth]{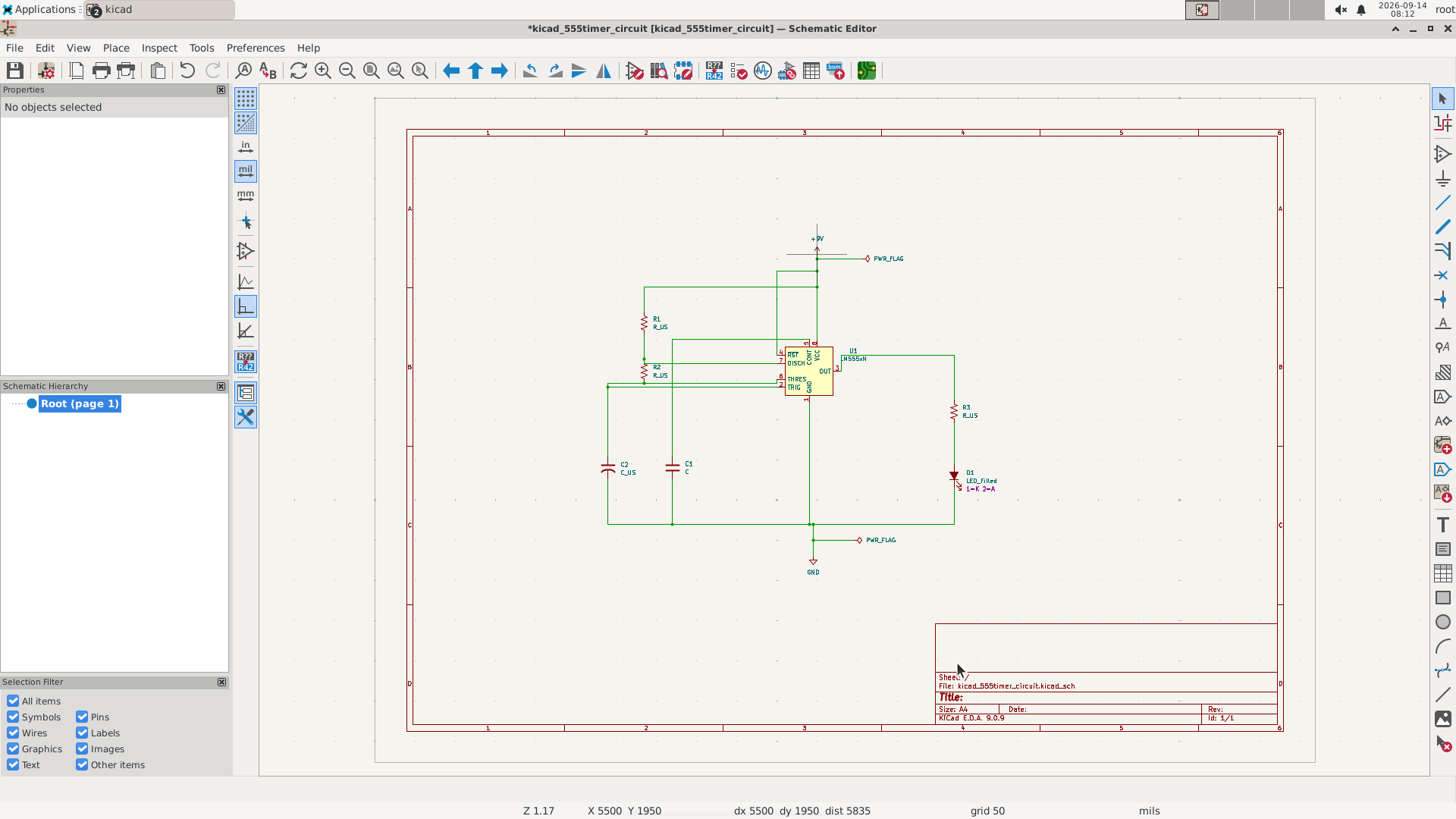}
    \caption{VISTA, 55f2eefb: label still reads +9\,V.}
    \label{fig:s010_label_still_reads_9v_on_screen_agent_mis}
  \end{subfigure}\hfill
  \begin{subfigure}[t]{0.32\textwidth}
    \centering
    \includegraphics[width=\linewidth]{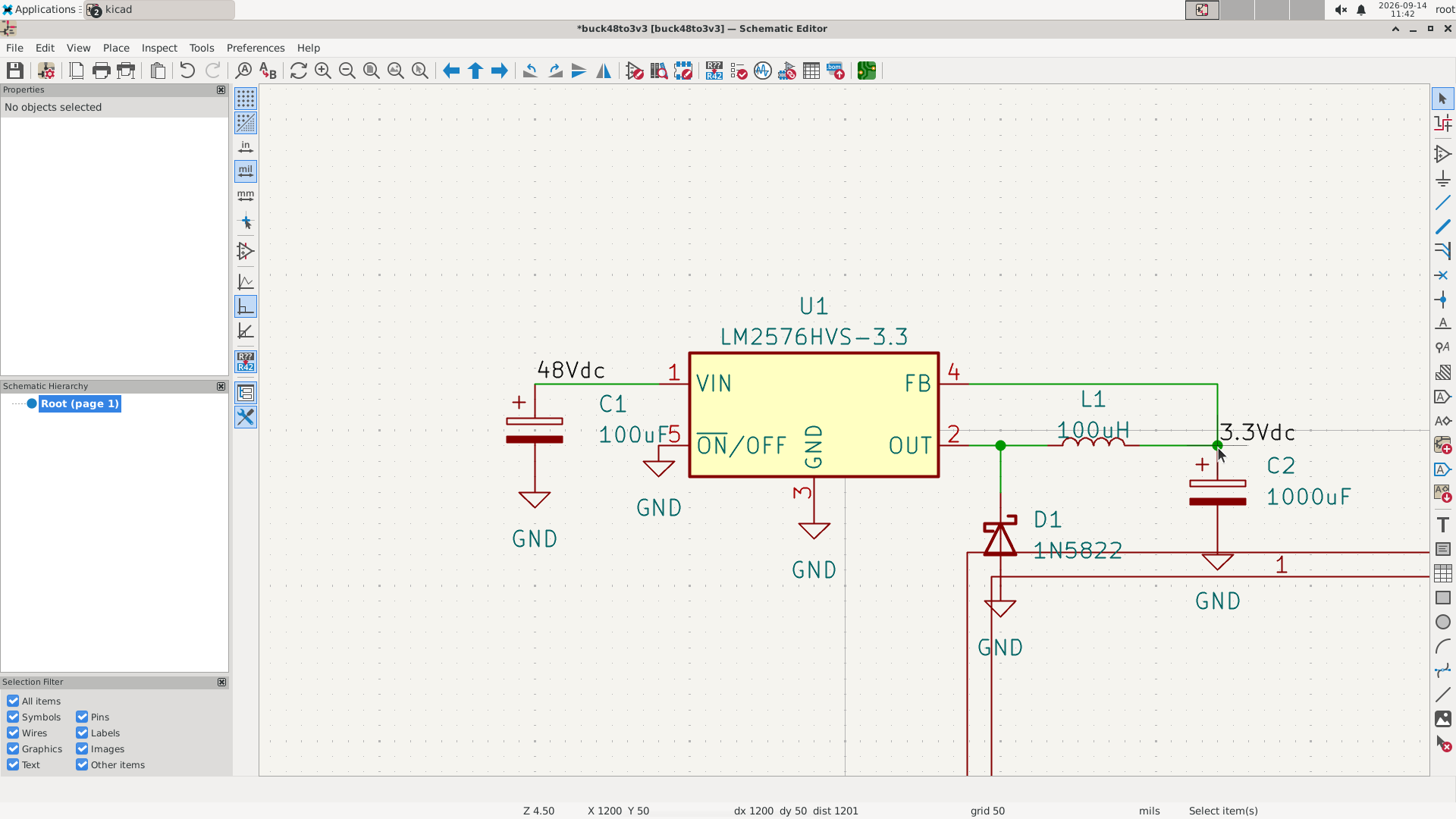}
    \caption{VISTA, c4baee36: local label \texttt{3.3VDC} on the output net.}
    \label{fig:s060_local_label_3_3vdc_on_the_output_net}
  \end{subfigure}\hfill
  \begin{subfigure}[t]{0.32\textwidth}
    \centering
    \includegraphics[width=\linewidth]{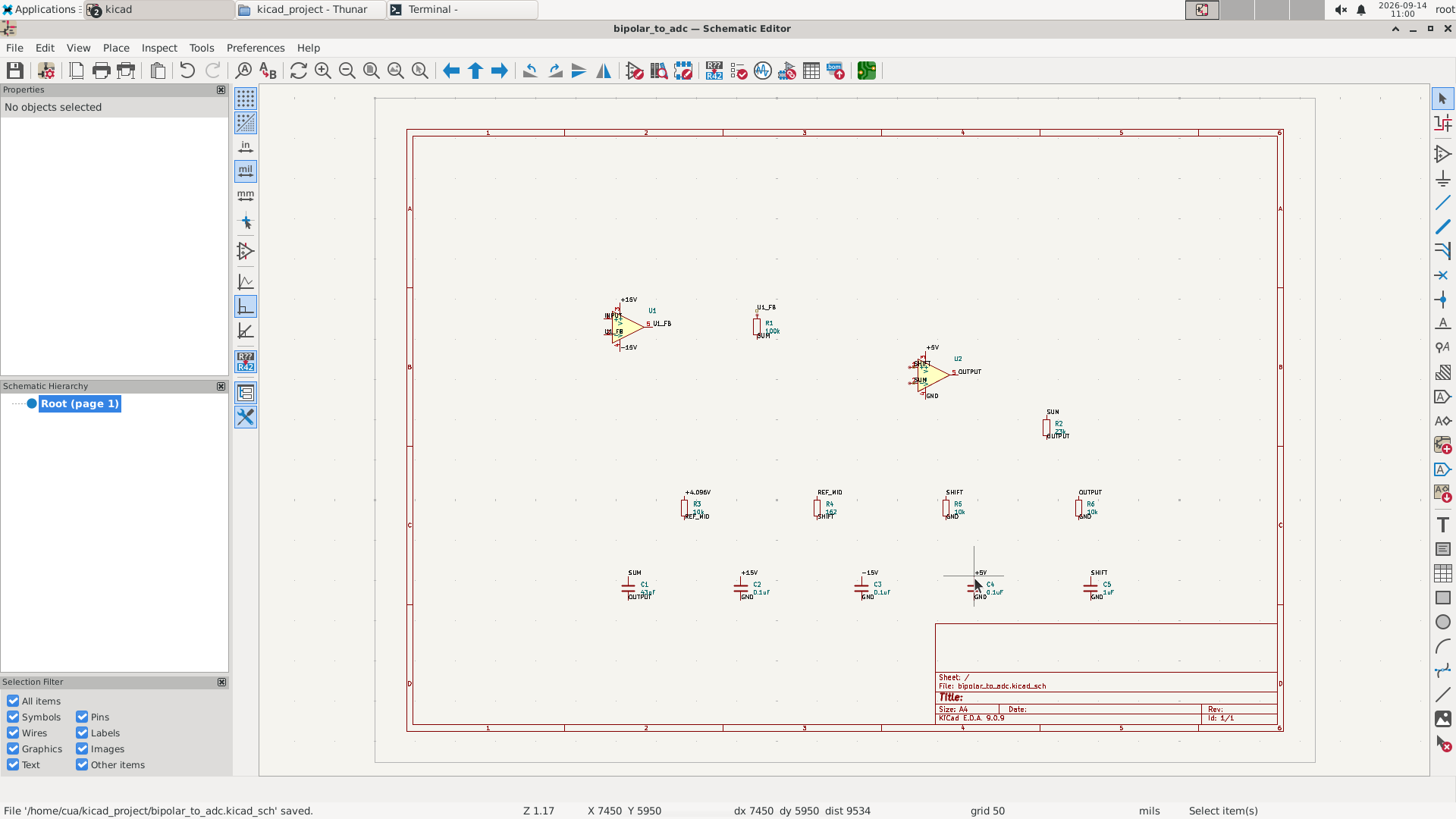}
    \caption{VISTA, b7148878: pins R1.1, U2.1, and U2.2 unconnected.}
    \label{fig:s125_saved_done_netlist_r1_1_u2_1_u2_2_dangli}
  \end{subfigure}
  \caption{\textbf{Qualitative comparison on CUABench-KiCad.}
  Final schematics produced by MFM (top) and VISTA (bottom); each column is one task.
  (a,\,d)~Task 55f2eefb, which switches an LED circuit's supply from 9\,V to 5\,V: VISTA's label still reads +9\,V even though Change Symbol returned ``OK''.
  (b,\,e)~Task c4baee36, a buck regulator: both agents place the same components and wiring but use different supply labels.
  (c,\,f)~Task b7148878, a two-op-amp circuit: MFM leaves one pin unconnected, compared with three for VISTA.}
  \label{fig:kicad-tasks-different-perf}
\end{figure}

\Cref{fig:kicad-tasks-different-perf} shows the final schematics that MFM and VISTA produce on the three tasks discussed in \Cref{sec:qualitative}.
On Task 55f2eefb, both agents get an ``OK'' from Change Symbol, but only MFM checks the resulting value; VISTA's schematic still reads +9\,V. Because MFM checks the resulting application state instead of trusting the ``OK'' returned by an edit operation, it catches incomplete edits that VISTA records as successful.
On Task c4baee36, both agents place the same components and wiring but use different supply labels.
On Task b7148878, MFM leaves one pin unconnected, compared with three for VISTA. 

\begin{figure}[t]
  \centering
  \begin{subfigure}[t]{0.32\textwidth}
    \centering
    \includegraphics[width=\linewidth]{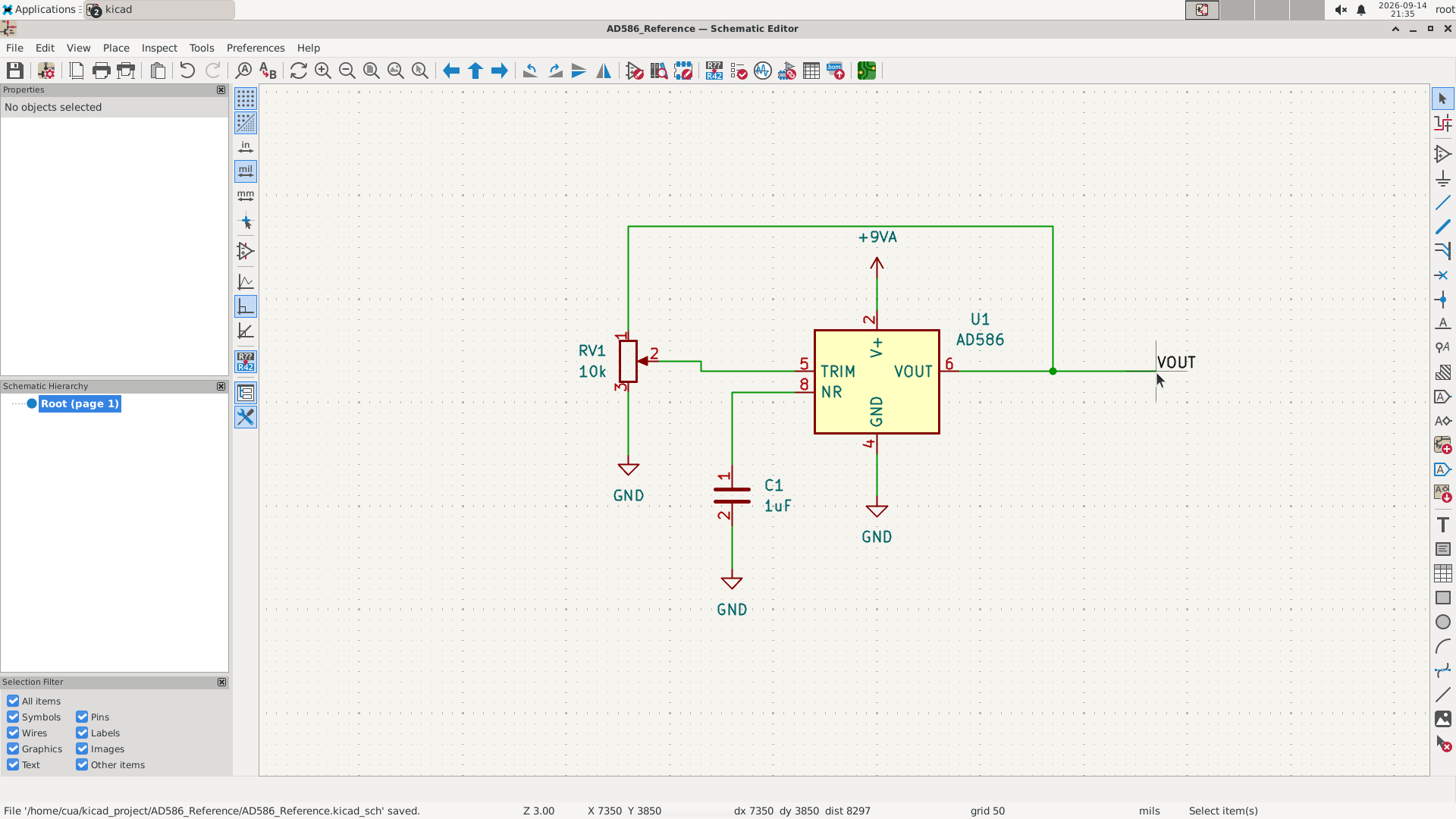}
    \caption{\texttt{ba34f1c3}: the label and all five nets match the design; the title bar confirms the save.}
    \label{fig:ba34f1c3_step_45}
  \end{subfigure}\hfill
  \begin{subfigure}[t]{0.32\textwidth}
    \centering
    \includegraphics[width=\linewidth]{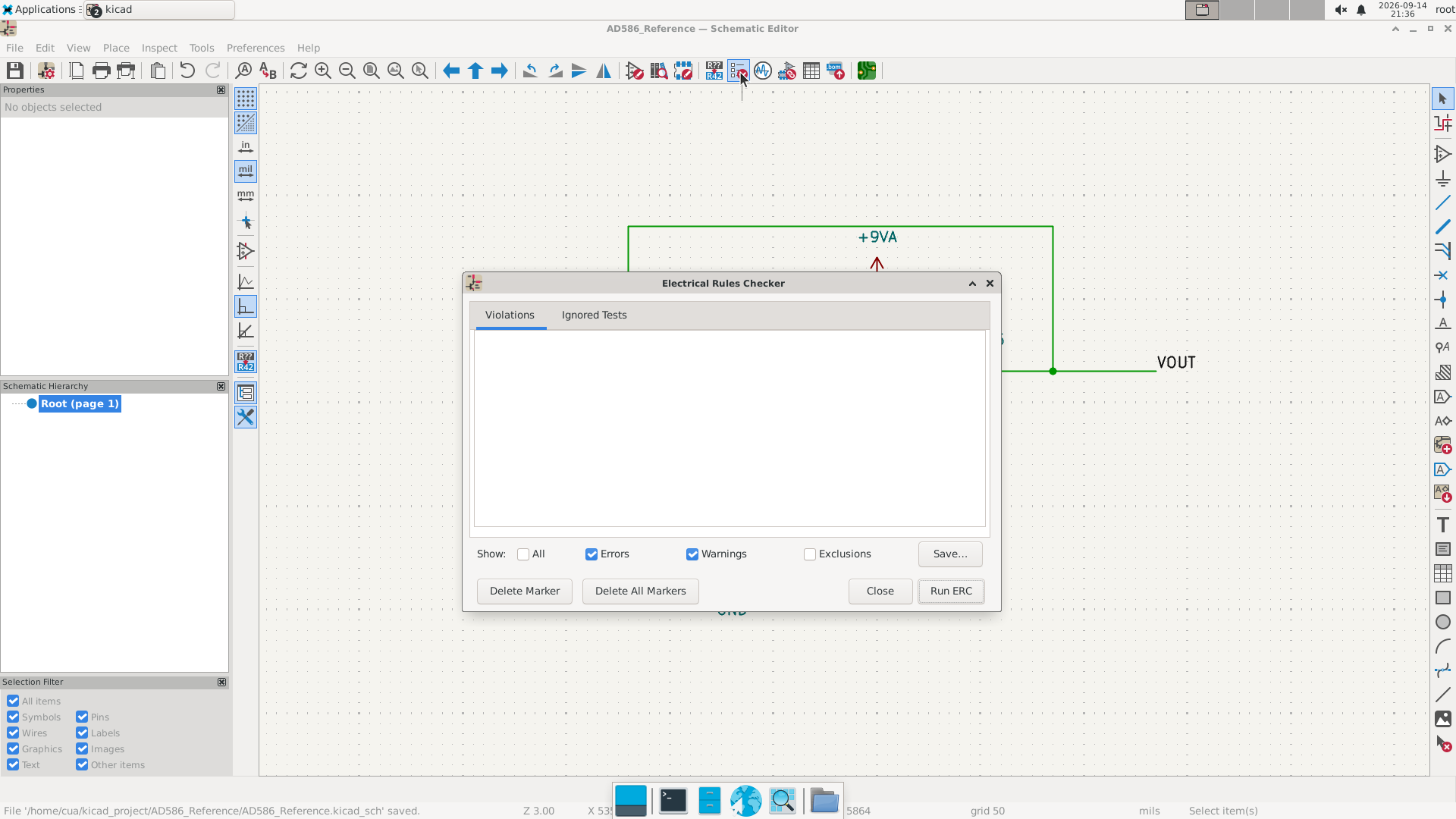}
    \caption{ERC dialog open; no error feedback from earlier actions.}
    \label{fig:ba34f1c3_step_46}
  \end{subfigure}\hfill
  \begin{subfigure}[t]{0.32\textwidth}
    \centering
    \includegraphics[width=\linewidth]{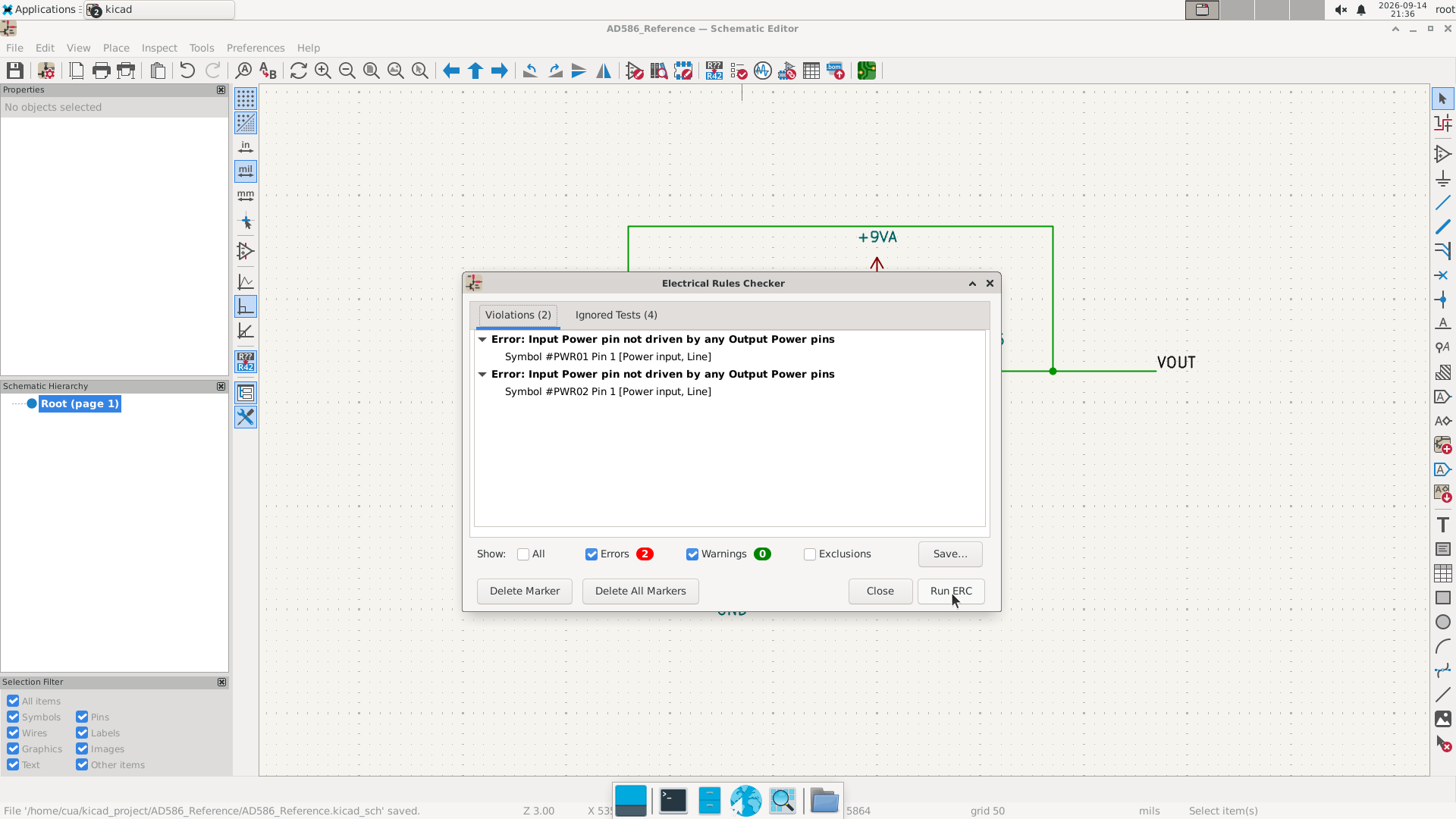}
    \caption{Only the two expected undriven-power errors; zero warnings.}
    \label{fig:ba34f1c3_step_47}
  \end{subfigure}

  \vspace{0.5em}
  \begin{subfigure}[t]{0.32\textwidth}
    \centering
    \includegraphics[width=\linewidth]{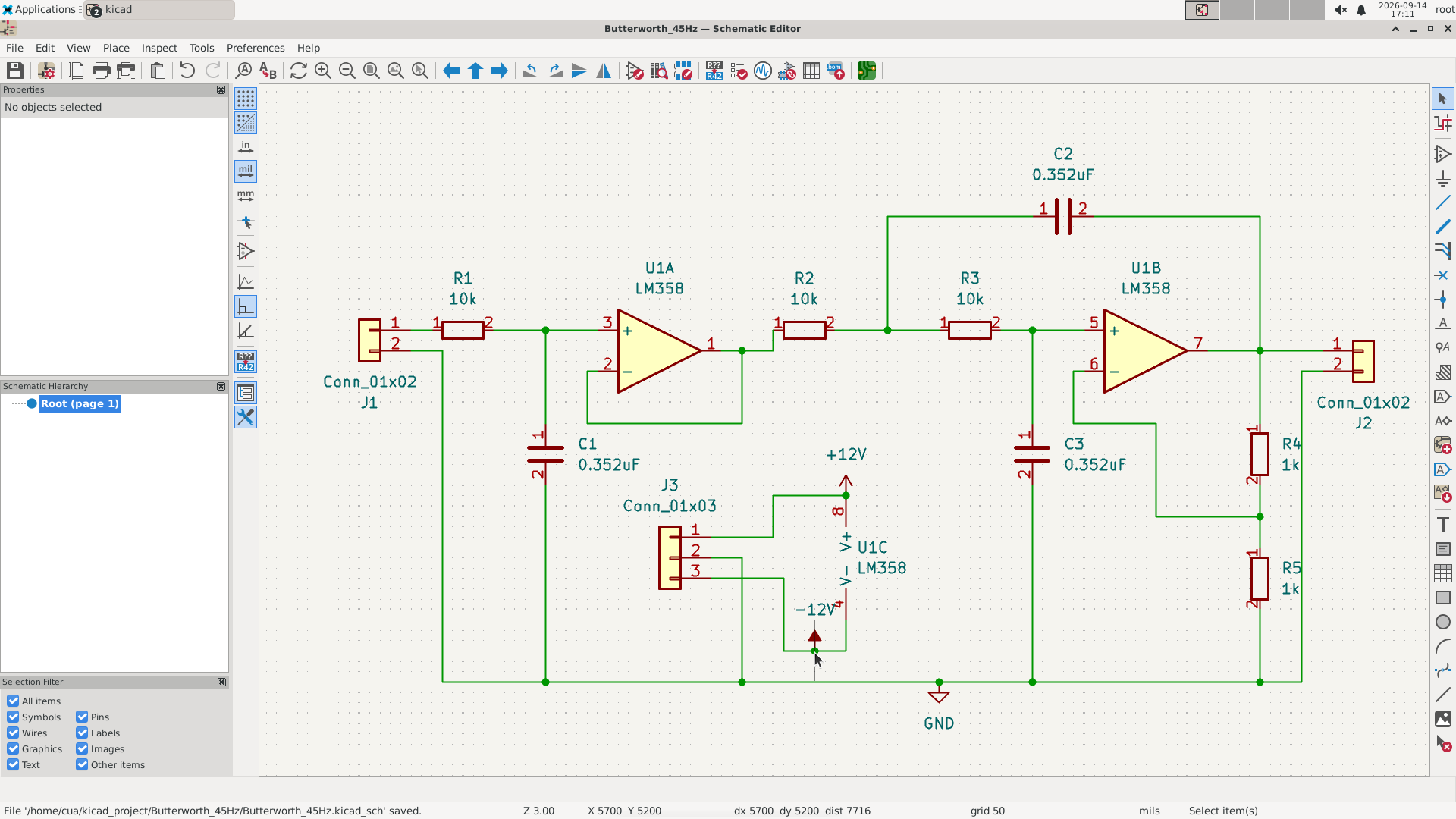}
    \caption{\texttt{4c913e30}: wiring matches the committed design; the schematic is saved in the required directory.}
    \label{fig:4c913e30_step_76}
  \end{subfigure}\hfill
  \begin{subfigure}[t]{0.32\textwidth}
    \centering
    \includegraphics[width=\linewidth]{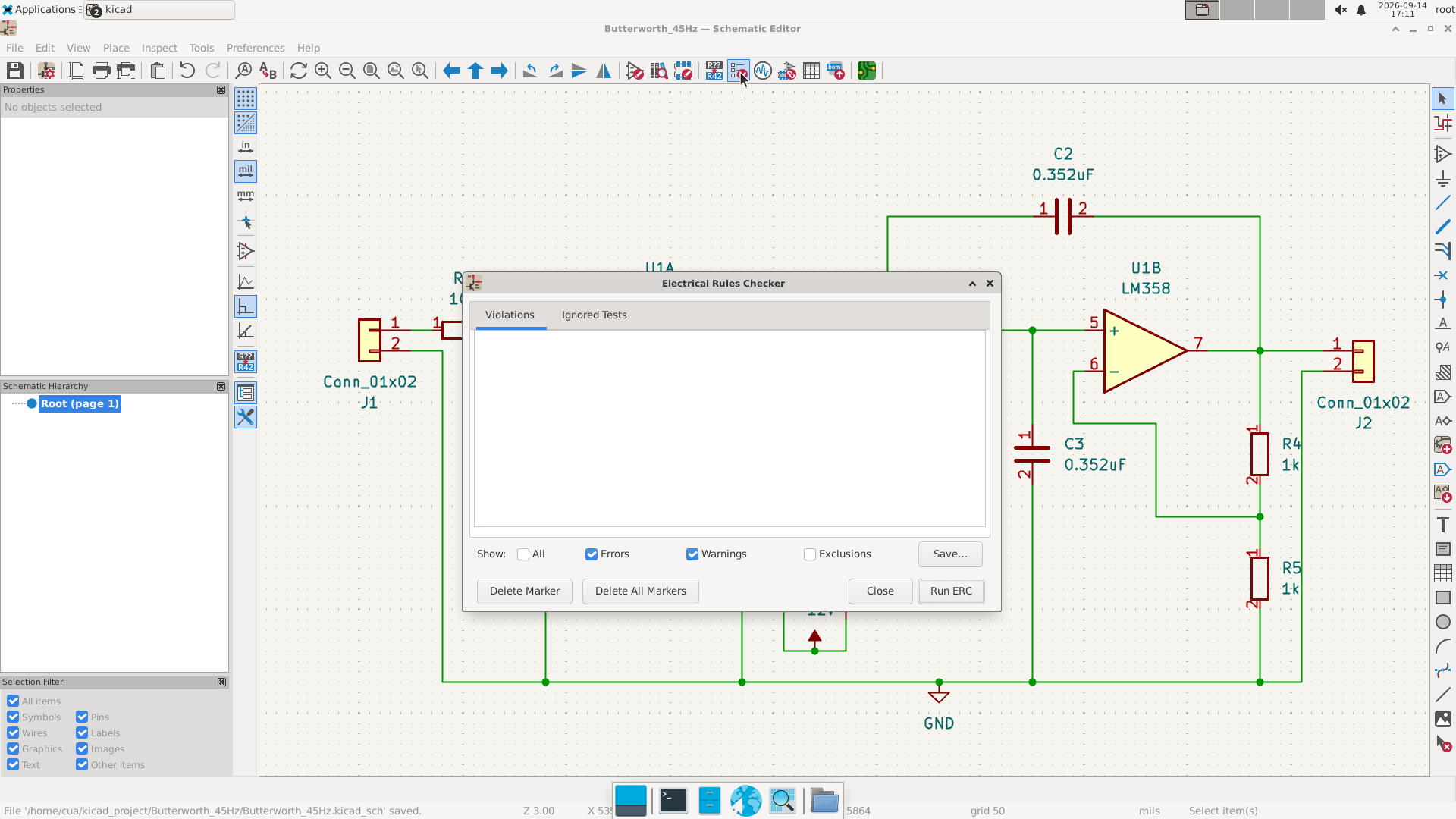}
    \caption{ERC dialog open; no error feedback from earlier actions.}
    \label{fig:4c913e30_step_77}
  \end{subfigure}\hfill
  \begin{subfigure}[t]{0.32\textwidth}
    \centering
    \includegraphics[width=\linewidth]{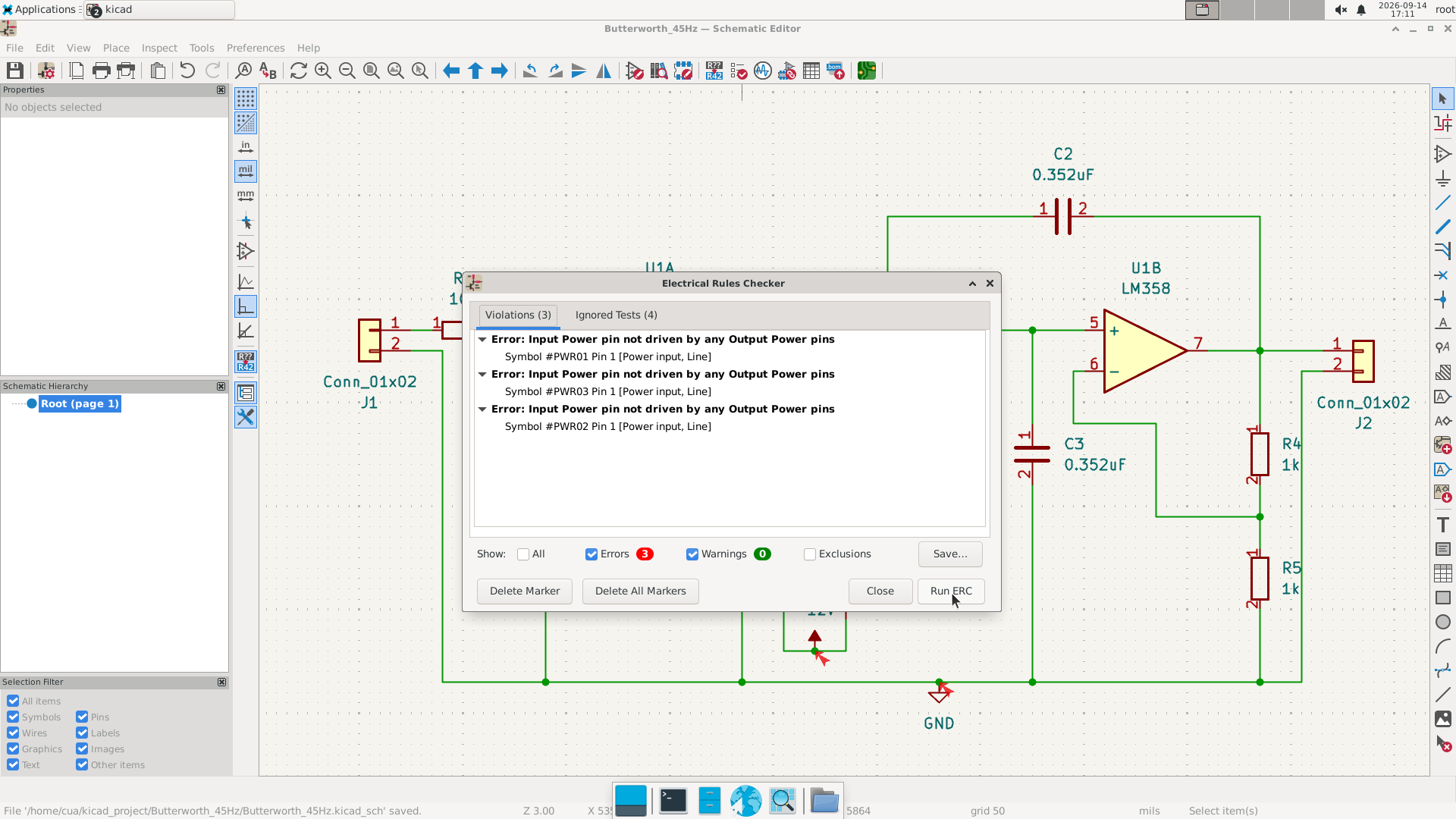}
    \caption{Only the three expected undriven-power errors (external supplies); zero warnings.}
    \label{fig:4c913e30_step_78}
  \end{subfigure}

  \vspace{0.5em}
  \begin{subfigure}[t]{0.32\textwidth}
    \centering
    \includegraphics[width=\linewidth]{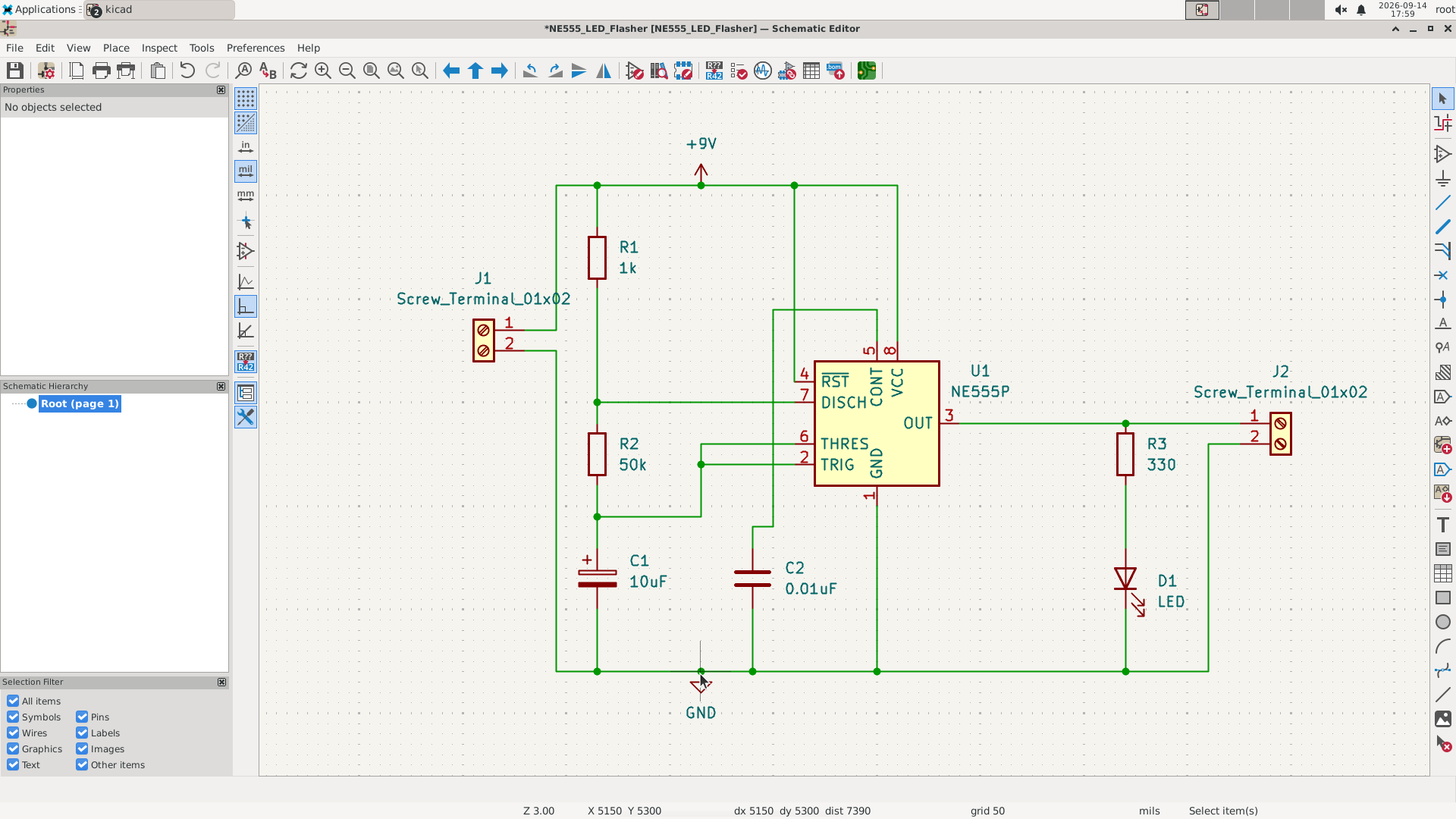}
    \caption{\texttt{5dbc9e02}: values, polarities, and all seven nets match the design.}
    \label{fig:5dbc9e02_step_68}
  \end{subfigure}\hfill
  \begin{subfigure}[t]{0.32\textwidth}
    \centering
    \includegraphics[width=\linewidth]{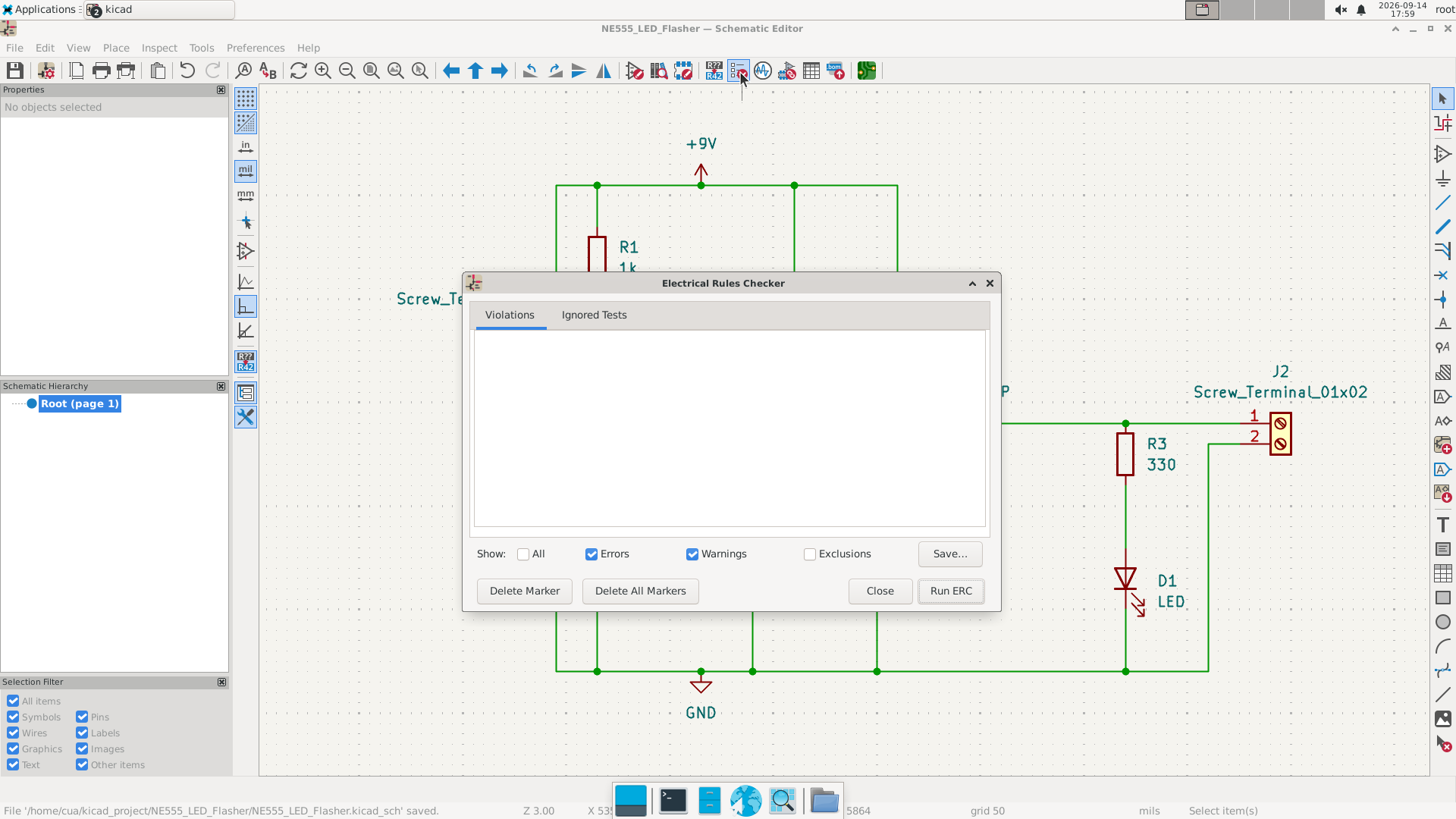}
    \caption{The title and status bar confirm the save location; ERC dialog open.}
    \label{fig:5dbc9e02_step_69}
  \end{subfigure}\hfill
  \begin{subfigure}[t]{0.32\textwidth}
    \centering
    \includegraphics[width=\linewidth]{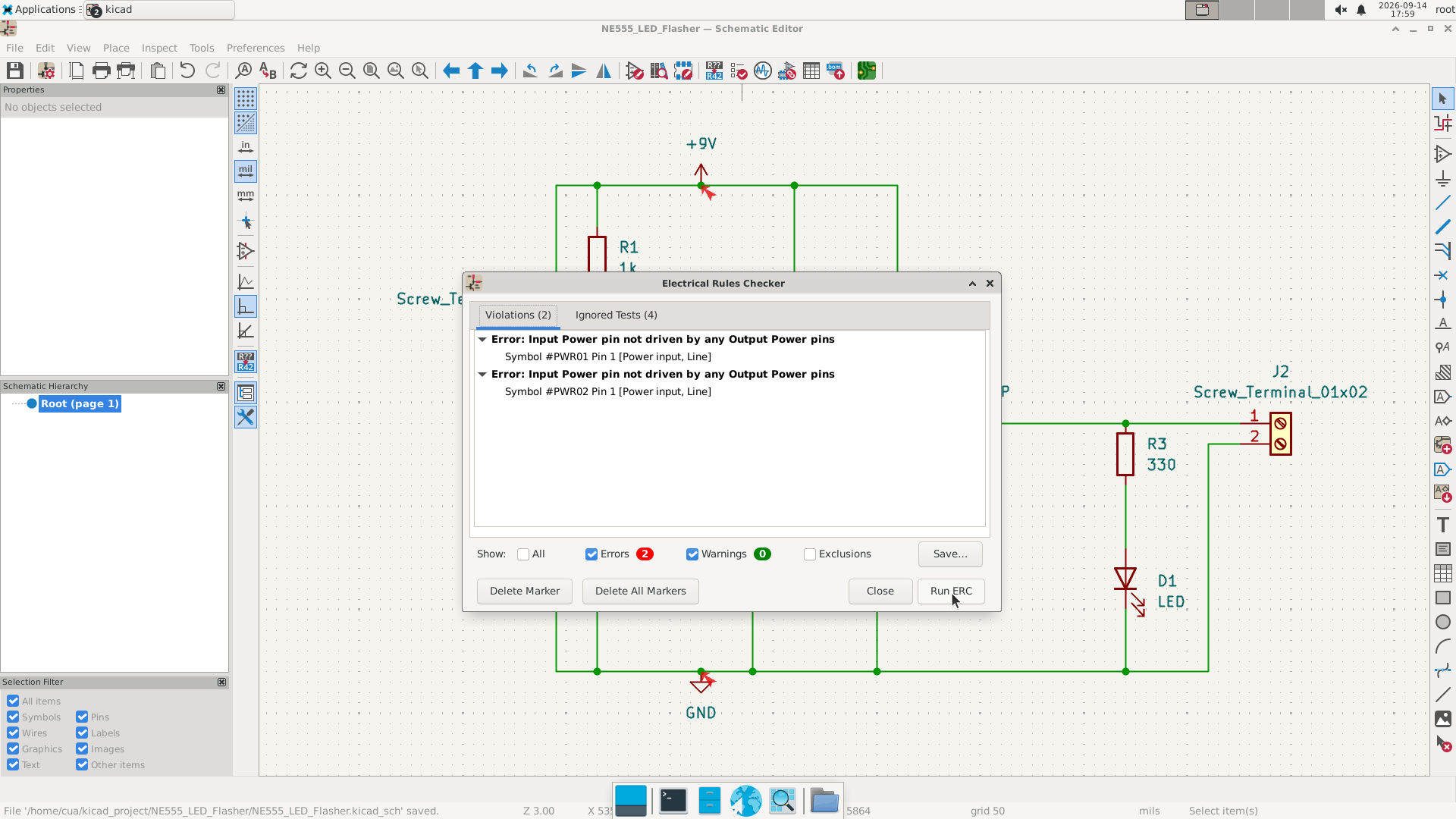}
    \caption{Two expected undriven-power errors; zero warnings and no unconnected pins.}
    \label{fig:5dbc9e02_step_70}
  \end{subfigure}
  \caption{\textbf{Self-verification of edits on CUABench-KiCad.}
  Each row shows one task.
  MFM inspects the schematic against the specification (left), opens the Electrical Rules Checker (middle), and reads the report to confirm that only the expected errors remain (right).}
  \label{fig:kicad-erc-verification}
\end{figure}

\begin{figure}[t]
  \centering
  \begin{subfigure}[t]{\textwidth}
    \centering
    \includegraphics[width=\linewidth]{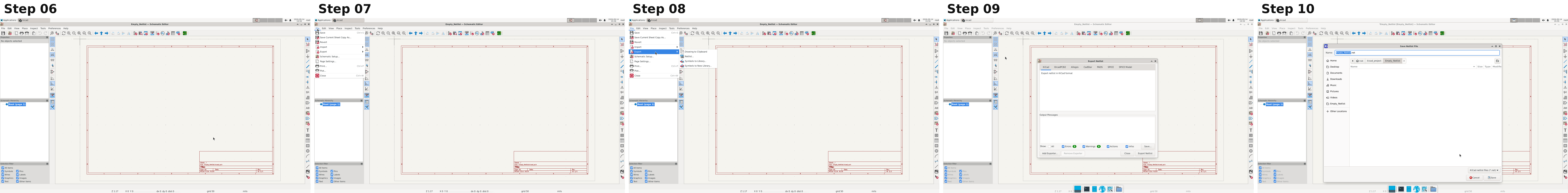}
    \caption{Current task \texttt{a6414tw4}, steps 6--10.}
    \label{fig:a6414tw4_steps_06-10}
  \end{subfigure}

  \vspace{0.5em}
  \begin{subfigure}[t]{\textwidth}
    \centering
    \includegraphics[width=\linewidth]{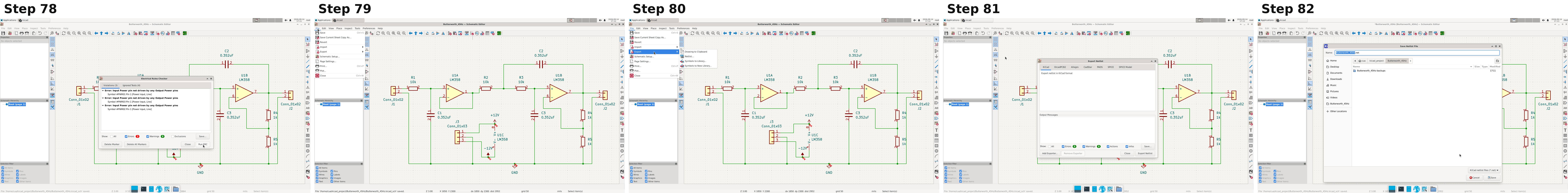}
    \caption{Episode retrieved from the earlier task \texttt{4c913e30}, steps 78--82.}
    \label{fig:4c913e30_steps_78-82}
  \end{subfigure}
  \caption{\textbf{Cross-task reuse of episodic memory in KiCad.}
  While working on the current task (top), the agent retrieves an episode from an earlier task (bottom) with a nearly identical action sequence.
  The sequence is: save the blank schematic, open \emph{Export}~$\rightarrow$~\emph{Netlist}, select the native KiCad format, and set the output path.
  The retrieved episode serves as a concrete precedent for the current task.}
  \label{fig:kicad-cross-task-reuse}
\end{figure}

\Cref{fig:kicad-erc-verification} shows how MFM verifies its own edits on three other tasks.
In each task, the agent follows the same three-step routine.
First, it inspects the schematic to confirm that component values and net labels match the specification.
Second, it runs the Electrical Rules Checker (ERC).
Third, it reads the ERC report to confirm that only the expected undriven-power errors remain and that no pins are left unconnected.
The agent first used this routine in an earlier task and later retrieved it from episodic memory, similar to the netlist-export reuse in \Cref{fig:kicad-cross-task-reuse}.

\Cref{fig:kicad-cross-task-reuse} shows episode reuse across tasks.
While exporting a netlist in Task \texttt{a6414tw4}, the agent retrieves an episode from the earlier Task \texttt{4c913e30} whose action sequence is nearly identical, and follows it as a precedent.

\subsection{Inspection Statistics} 
\label{app:inspection_stats}

\begin{figure}[ht]
\centering
\includegraphics[width=\textwidth]{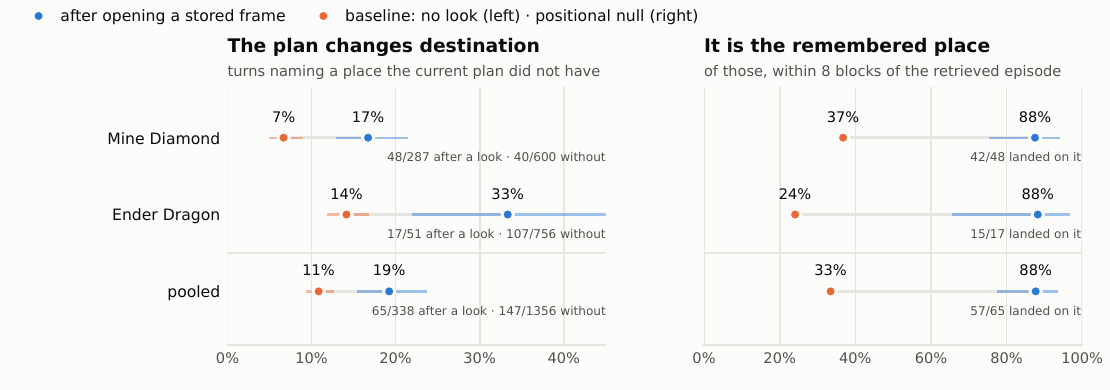}
\caption{ \textbf{What follows a look at a stored frame.} \emph{Left:} the share of turns whose
plan names a destination the current plan did not have, after the agent opened a stored
episode frame (blue) against turns in which it opened no image at all (orange).
\emph{Right:} of those new destinations, the share falling within $8$ blocks of the
position at which the retrieved episode was recorded (blue), against the share of the
whole episode store that would match the same coordinate by position alone (orange).}
\label{fig:frame-look}
\end{figure}

We quantify a mechanism behind Sec.~\ref{sec:qualitative} over the main Minecraft runs (three models, two tasks, three seeds): how often inspecting a stored frame changed the agent's next action (Fig.~\ref{fig:frame-look}). 

\textbf{Does looking at a stored frame change the next action?}
The agent looks \emph{before} it commits, and the transcripts hold no counterfactual, so a
within-call before/after comparison is empty by construction. What can be measured is the
destination. A turn is a \emph{retarget} if its plan names a
\texttt{go\_to}/\texttt{look\_at} coordinate the running plan did not contain, and a
\emph{recall target} if that coordinate lies within $8$ blocks of where the looked-at
episode was recorded; the null for recall is the fraction of all stored episodes that
would match the coordinate by position alone. On Mine Diamond
(\Cref{fig:frame-look}), a turn that opened a stored frame retargets $2.5\times$ as often
as one that did not ($17\%$ vs $7\%$, $p=5\times10^{-6}$, Fisher), and the new destination
is the retrieved episode's location in $42$ of $48$ cases against a $37\%$ null
($p=6\times10^{-13}$, binomial). Ender Dragon shows the same pattern ($33\%$ vs $14\%$;
$15/17$ against a $24\%$ null), as does the pool ($19\%$ vs $11\%$; $57/65$ against
$33\%$). The
retrieval is therefore not merely correlated with replanning: it selects \emph{which}
place the agent goes to next.

\clearpage

\section{Pseudocode}


\textbf{Notation.}
Let $g$ denote the task objective, $o_t=(v_t,x_t)$ an observation,
and $M_t=(E_t,I_t,C_t)$ the memory at step $t$.
Here $E_t$ stores completed multimodal episodes, $I_t$ provides searchable
cues and references to those episodes, and $C_t$ contains revisable
agent-written knowledge. The algorithms use $E$, $I$, and $C$ for their
current values. The initial memory may be empty or contain retained
experience. An episode groups a contiguous interaction interval;
$B$ holds the interval currently being recorded, and $\Vert$ denotes
sequence concatenation.

\begin{algorithm}[ht]
\caption{MFM: Multimodal Filesystem Memory}
\label{alg:mfm-main}
\begin{algorithmic}[1]
\Require Objective $g$; environment $\mathcal{E}$; initial memory $M$; horizon $T$
\Ensure Updated memory $M$
\State $(E,I,C)\gets M$; $B\gets\langle\rangle$; $o_1\gets\mathcal{E}.\textsc{Observe}()$
\For{$t=1,\ldots,T$}
    \State $D_t\gets\emptyset$
    \If{past experience is needed}
        \State $q_t\gets\Call{FormQuery}{g,C,\{o_t\}}$
        \State $D_t\gets\Call{Retrieve}{q_t,I}$
    \EndIf
    \If{consolidation is triggered}
        \State $C\gets\Call{Consolidate}{g,C,E,I,\{o_t\}\cup D_t}$
    \EndIf
    \State $a_t\gets\Call{SelectAction}{g,o_t,C,D_t}$
    \State $(o_{t+1},f_t)\gets\mathcal{E}.\textsc{Step}(a_t)$
    \State Persist $(o_t,a_t,o_{t+1},f_t)$ with their interaction identifiers
    \State Append references to these records to $B$
    \If{an episode boundary is reached}
        \State $(E,I)\gets\Call{RecordEpisode}{B,E,I}$; $B\gets\langle\rangle$
    \EndIf
    \If{execution terminates}
        \State \textbf{break}
    \EndIf
\EndFor
\If{$B\neq\langle\rangle$}
    \State $(E,I)\gets\Call{RecordEpisode}{B,E,I}$
\EndIf
\State \Return $M=(E,I,C)$
\end{algorithmic}
\end{algorithm}

\noindent
Here $f_t$ denotes available interaction feedback. Episode boundaries and
consolidation triggers are supplied by the interaction protocol. Retrieval
may be repeated as needed, and action selection may reuse an existing plan
or procedure. Each resulting observation is recorded before its episode
is finalized, including at termination or the end of the horizon.

\begin{algorithm}[ht]
\caption{Episodic recording and indexing}
\label{alg:mfm-record}
\begin{algorithmic}[1]
\Function{RecordEpisode}{$B,E,I$}
    \State $\mathit{id}_i\gets$ a unique episode identifier
    \State $(\tau_i,V_i,S_i,A_i,m_i)\gets\Call{OrganizeRecords}{B}$
    \State $e_i\gets(\mathit{id}_i,\tau_i,V_i,S_i,A_i,m_i)$
    \State $\rho_i\gets\Call{StoreEpisode}{e_i}$ \Comment{Preserve the original evidence}
    \State $k_i\gets\Call{BuildCues}{e_i}$
    \State $z_i\gets(\mathit{id}_i,k_i,\rho_i)$
    \State \Return $(E\Vert\langle e_i\rangle,\ I\Vert\langle z_i\rangle)$
\EndFunction
\end{algorithmic}
\end{algorithm}

\noindent
The episode contains its interval $\tau_i$, visual observations $V_i$,
structured states $S_i$, attempted actions $A_i$, and metadata $m_i$.
The cues $k_i$ include available temporal, spatial, action, and outcome
descriptions; $\rho_i$ references the underlying stored evidence.
Organizing and indexing records does not replace or compress their
original multimodal content.

\begin{algorithm}[t]
\caption{Retrieval through textual cues and evidence references}
\label{alg:mfm-retrieve}
\begin{algorithmic}[1]
\Function{Retrieve}{$q,I$}
    \State $R\gets\Call{Filter}{I,q}$ \Comment{Retain entries satisfying all query predicates}
    \State $P\gets\Call{SelectReferences}{q,R}$
    \State $D\gets\{\Call{Load}{p}:p\in P\}$
    \State \Return $D$
\EndFunction
\end{algorithmic}
\end{algorithm}

\noindent
Filtering preserves index order and may restrict results by recency when
requested. The agent selects $P\subseteq\bigcup_{z_i\in R}\rho_i$, specifying
which referenced observations or records to inspect. Textual cues locate
candidate episodes; the underlying evidence can contain details absent
from those cues.

\begin{algorithm}[t]
\caption{Self-evolving consolidation}
\label{alg:mfm-consolidate}
\begin{algorithmic}[1]
\Function{Consolidate}{$g,C,E,I,D$}
    \State $D^{\mathrm{cons}}\gets D$
    \While{additional evidence is needed}
        \State $q\gets\Call{FormQuery}{g,C,D^{\mathrm{cons}}}$
        \State $D^{\mathrm{cons}}\gets D^{\mathrm{cons}}\cup\Call{Retrieve}{q,I}$
    \EndWhile
    \State $C^+\gets\mathcal{U}(C,D^{\mathrm{cons}},g)$
    \State \Return $C^+$ \Comment{$E$ and $I$ remain unchanged}
\EndFunction
\end{algorithmic}
\end{algorithm}

\noindent
The agent's update $\mathcal{U}$ revises plans, findings, relational memory,
general guidance, or reusable procedures as appropriate. Relations may be
expressed as $\ell=(u,r,v;P_\ell)$, where $P_\ell$ references
supporting evidence. Consolidation changes interpretations while preserving
the episode archive and index; it does not execute environment actions.

\clearpage

\section{Skills}
\label{sec:app_skills}


\begin{myskillbox}[skill:cut-step]{cut step}{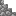}

"""cut_step: cut one descending step and advance; requires a pick, close solid dirt/stone wall, centered cardinal alignment, safe footing and no opening or GUI. Proven at steps5068-5249."""
import argparse,json,sys
PARAMS={"yaw":"cardinal heading toward the inspected solid wall"}
p=argparse.ArgumentParser();p.add_argument('--args',default='{}')
a=json.loads(p.parse_args().args);s=json.load(sys.stdin);acts=[]
if not s.get('gui_open'):
    yaw=a.get('yaw',s['pos']['yaw'])
    for pitch in (0,45,65):
        acts += [{'procedure':'look_abs','args':{'pitch':pitch,'yaw':yaw,'ticks':10}},{'action':{'attack':1},'repeat':45}]
    acts += [{'action':{'forward':1},'repeat':8},{'action':{},'repeat':8},{'procedure':'look_abs','args':{'pitch':60,'yaw':yaw,'ticks':10}}]
print(json.dumps({'actions':acts}))

\end{myskillbox}

\begin{myskillbox}[skill:quarry-pair]{quarry pair}{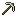}

"""quarry_north_pair: mine one upper/lower north wall pair and collect; requires suitable pick held, solid close stone wall north, level solid floor, no GUI or opening; sweep advances about one block."""
import json,sys
s=json.load(sys.stdin)
a=[{'procedure': 'look_abs', 'args': {'pitch': 0, 'yaw': 180, 'ticks': 10}}, {'action': {'attack': 1}, 'repeat': 30}, {'procedure': 'look_abs', 'args': {'pitch': 50, 'yaw': 180, 'ticks': 10}}, {'action': {'attack': 1}, 'repeat': 30}, {'procedure': 'collect_drops', 'args': {'ticks': 24}}] if not s.get('gui_open') else []
print(json.dumps({'actions':a}))

\end{myskillbox}

\begin{myskillbox}[skill:bridge-north]{bridge north}{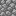}

"""bridge_north: extend one northern floor block using continuous sneak; face within 90 degrees of north with cobblestone held, clear headroom, and verified floor ahead. Repaired after two raw successes."""
import argparse
import json
import math
import sys
PARAMS = {"x": "bridge floor x", "y": "bridge floor y", "z": "new northern floor cell z"}
p = argparse.ArgumentParser()
p.add_argument('--args', default='{}')
a = json.loads(p.parse_args().args)
s = json.load(sys.stdin)
pos = s['pos']
actions = []
north_error = abs((pos['yaw'] - 180 + 180) % 360 - 180)
if (not s.get('gui_open') and s.get('inventory', {}).get('cobblestone', 0) > 0
    and north_error <= 90 and -10 <= pos['pitch'] <= 80
    and math.floor(pos['x']) == a['x'] and abs(pos['y'] - a['y'] - 1) < .05
    and math.floor(pos['z']) - 1 == a['z']):
    actions = [
        {'action': {'camera': [80-pos['pitch'], (180-pos['yaw']+180)\%360-180], 'sneak': 1}, 'repeat': 1},
        {'action': {'forward': 1, 'sneak': 1}, 'repeat': 20},
        {'action': {'sneak': 1}, 'repeat': 8},
        {'action': {'camera': [1, -90], 'sneak': 1}, 'repeat': 1},
        {'action': {'camera': [0, -90], 'sneak': 1}, 'repeat': 1},
        {'action': {'use': 1, 'sneak': 1}, 'repeat': 1},
        {'action': {'sneak': 1}, 'repeat': 8},
        {'action': {'camera': [-16, 90], 'sneak': 1}, 'repeat': 1},
        {'action': {'camera': [0, 90], 'sneak': 1}, 'repeat': 1},
        {'action': {'sneak': 1}, 'repeat': 8}
    ]
print(json.dumps({'actions': actions}))

\end{myskillbox}

\begin{myskillbox}[skill:shaft-down]{Shaft down}{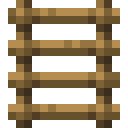}
import json
import math
import sys

PARAMS = {"target_y": "target feet height, minimum 12", "pick": "pickaxe hotbar slot 1-9", "max_steps": "mining budget, maximum 370"}
arguments = json.loads(sys.argv[sys.argv.index("--args") + 1]) if "--args" in sys.argv else {}
state = json.load(sys.stdin)
position = state.get("pos")
if not position or state.get("gui_open"):
    raise SystemExit("close the screen and stand on solid ground first")
if abs(position["y"] - round(position["y"])) > 0.05:
    raise SystemExit("not standing at an integer block height")
for coordinate in ("x", "z"):
    fraction = position[coordinate] - math.floor(position[coordinate])
    if not 0.30 <= fraction <= 0.70:
        raise SystemExit("center within the block before descending")
speeds = {"stone_pickaxe": (22, 19), "iron_pickaxe": (18, 15), "diamond_pickaxe": (15, 12)}
pickaxes = [entry for entry in state["hotbar"] if entry.get("item") in speeds]
if "pick" in arguments:
    pickaxes = [entry for entry in pickaxes if int(entry["slot"]) + 1 == int(arguments["pick"])]
if not pickaxes:
    raise SystemExit("requested pickaxe is not in the hotbar")
pickaxe = min(pickaxes, key=lambda entry: speeds[entry["item"]][0])
height = int(round(position["y"]))
target = max(12, int(arguments.get("target_y", 12)))
if height <= target:
    raise SystemExit("target depth reached; mine horizontally")
normal_speed, minimum_speed = speeds[pickaxe["item"]]
steps = min(370, max(1, int(arguments.get("max_steps", 300))), (height - target) * normal_speed, (height - 11) * minimum_speed)
print(json.dumps({"actions": [
    {"procedure": "equip_hotbar", "args": {"slot": int(pickaxe["slot"]) + 1}},
    {"procedure": "look_abs", "args": {"pitch": 88, "steps": 8}},
    {"procedure": "mine_ahead", "args": {"n": steps, "ore_watch": True}},
    {"procedure": "look_abs", "args": {"pitch": 25, "steps": 8}}
]}))
\end{myskillbox}

\begin{myskillbox}[skill:tunnel]{Tunnel}{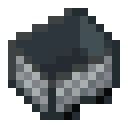}
"""tunnel: extend a level two-block-high tunnel by up to four cells, watching for ore and hazards."""
import json
import sys

PARAMS = {"cells": "one to four cells", "yaw": "cardinal tunnel heading", "pick": "pickaxe hotbar slot 1-9"}
arguments = json.loads(sys.argv[sys.argv.index("--args") + 1]) if "--args" in sys.argv else {}
state = json.load(sys.stdin)
position = state.get("pos")
if not position or state.get("gui_open"):
    raise SystemExit("close the screen before tunneling")
if abs(position["y"] - round(position["y"])) > 0.05:
    raise SystemExit("stand on solid ground before tunneling")
if position["y"] < 11:
    raise SystemExit("restore feet to y11 or higher before tunneling")
heading = float(arguments.get("yaw", position["yaw"]))
if abs(heading / 90 - round(heading / 90)) > 0.01:
    raise SystemExit("choose a cardinal heading")
slot = int(arguments.get("pick", 7))
pickaxes = [entry for entry in state["hotbar"] if int(entry["slot"]) + 1 == slot and entry.get("item") in ("iron_pickaxe", "stone_pickaxe", "diamond_pickaxe")]
if not pickaxes:
    raise SystemExit("requested pickaxe is missing")
cells = min(4, max(1, int(arguments.get("cells", 4))))
actions = [{"procedure": "equip_hotbar", "args": {"slot": slot}}]
for cell in range(cells):
    actions.extend([
        {"procedure": "look_abs", "args": {"pitch": 0, "yaw": heading, "steps": 8}},
        {"action": {"attack": 1}, "repeat": 30},
        {"procedure": "look_abs", "args": {"pitch": 65, "yaw": heading, "steps": 8}},
        {"action": {"attack": 1}, "repeat": 30},
        {"action": {"forward": 1}, "repeat": 12}
    ])
actions.append({"procedure": "look_abs", "args": {"pitch": 0, "yaw": heading, "steps": 8}})
print(json.dumps({"actions": actions}))

\end{myskillbox}

\begin{myskillbox}[skill:probe]{Probe}{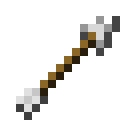}
"""probe: mine a stationary three-block side probe at a gentle downward angle, watching for ore."""
import json
import sys

PARAMS = {"yaw": "cardinal probe heading", "pick": "pickaxe hotbar slot 1-9"}
arguments = json.loads(sys.argv[sys.argv.index("--args") + 1]) if "--args" in sys.argv else {}
state = json.load(sys.stdin)
position = state.get("pos")
if not position or state.get("gui_open"):
    raise SystemExit("close the screen before probing")
if position["y"] < 11 or abs(position["y"] - round(position["y"])) > 0.05:
    raise SystemExit("stand on solid ground at y11 or higher before probing")
heading = float(arguments.get("yaw", position["yaw"]))
if abs(heading / 90 - round(heading / 90)) > 0.01:
    raise SystemExit("choose a cardinal heading")
slot = int(arguments.get("pick", 8))
pickaxes = [entry for entry in state["hotbar"] if int(entry["slot"]) + 1 == slot and entry.get("item") in ("iron_pickaxe", "stone_pickaxe", "diamond_pickaxe")]
if not pickaxes:
    raise SystemExit("requested pickaxe is missing")
actions = [
    {"procedure": "equip_hotbar", "args": {"slot": slot}},
    {"procedure": "look_abs", "args": {"pitch": 15, "yaw": heading, "steps": 8}}
]
for block in range(3):
    actions.extend([
        {"action": {"attack": 1}, "repeat": 30},
        {"action": {}, "repeat": 1}
    ])
print(json.dumps({"actions": actions}))

\end{myskillbox}

\begin{myskillbox}[skill:spruce-approach]{Spruce Approach}{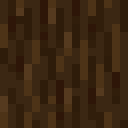}
First time:
```
Action 1 | Step 1     [ACTION] look(pitch=0,yaw=18) -> 1 ticks
Action 2 | Step 26    [ACTION] {"attack":1,"forward":1} x25
                      [STATE] pos=(-3010.70, 71.00, -5569.30) pitch=0 yaw=18 moved=3.42
                      [EVENT] mine_block spruce_leaves +1  (x2)
Action 3 | Step 226   [ACTION] mine_ahead(n=200) -> 200 ticks
                      [EVENT] mine_block spruce_leaves +1  (x4)
                      [NOTE] mine_ahead: finished 200 ticks, broke 4 blocks.
Action 4 | Step 227   [PLAN] OBSERVED vs EXPECTED: Leaves broke, but no wood; the exposed trunk is left of the crosshair.
                      [ACTION] look(yaw=-28,pitch=0) -> 1 ticks
Action 5 | Step 297   [ACTION] mine_ahead(n=100) -> 100 ticks
                      [EVENT] MILESTONE mine_log verified
                      [EVENT] mine_block spruce_log +1
```

Second time:
```json
{"actions":[{"procedure":"look_abs","args":{"pitch":0,"yaw":18,"steps":8}},
            {"action":{"attack":1,"forward":1},"repeat":25},
            {"procedure":"mine_ahead","args":{"n":70}},
            {"procedure":"look_abs","args":{"pitch":0,"yaw":-10,"steps":8}},
            {"procedure":"mine_ahead","args":{"n":150}}]

}
```


Third time:
> Spawn spruce approach verified two times: yaw18, attack+forward25, mine_ahead70, yaw-10, mine150.

\end{myskillbox}

\begin{myskillbox}[skill:blaze-fight]{blaze\_fight}{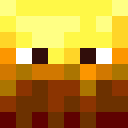}

The detector:
```
def find_blazes(path, ymax=150):
    """Return blobs [{'x','y','area','h','dist'}], nearest (biggest) first."""
    from PIL import Image
    im = Image.open(path).convert("RGB")
    if im.size != (W, H):
        im = im.resize((W, H))
    px = im.load()
    # A BLAZE IS YELLOW, FIRE IS ORANGE: the burning grass and the flames on the trees here
    # measure (165, 66, 5) -- green 66 -- while a blaze's rods are (167, 167, 79). The warm
    # mask takes both (a blaze close up is wrapped in its own orange flames); `yellows` marks
    # the rod-coloured ones, and a group must contain at least one to count as a mob.
    warm = set()
    yellows = set()
    for y in range(0, ymax):
        row_guard = y < 40
        for x in range(0, W):
            if row_guard and x >= 296:
                continue
            # my own arm holding the sword is screen-fixed at the bottom right and reads warm
            if x >= 282 and y >= 42:
                continue
            r, g, b = px[x, y]
            if r > 140 and g > 80 and b < 130 and r - b > 60 and g - b > 20:
                warm.add((x, y))
                if g > 110 and r - g < 70 and g - b > 40:
                    yellows.add((x, y))
    ...  # erode on the 4-neighbour cross, then merge fragments on a 13-px grid (union-find)
    for g, yel in merged.values():
        if len(g) < MIN_AREA:
            continue
        # fire is orange THROUGHOUT; a blaze always shows its yellow rods ... every real
        # blaze frame measured here runs 33\% to 100\%. A 10\% line separates them.
        if yel < YELLOW or yel < 0.10 * len(g):
            continue
        xs = [q[0] for q in g]
        ys = [q[1] for q in g]
        h = float(max(ys) - min(ys) + 1)
        ...
        out.append({"x": sum(xs) / float(len(xs)), "y": sum(ys) / float(len(ys)),
                    "area": float(len(g)), "h": h, "yel": float(yel), "dtop": float(dtop),
                    "w": float(max(xs) - min(xs) + 1),
                    "dist": BLAZE_H * FOCAL / max(2.0, h)})
    out.sort(key=lambda b: -b["area"])
    return out

```

Aim and swing:
```
    dyaw = _cap(_deg(tgt["x"] - CX))
    tdown = MAXDOWN_Y if tgt["yel"] / max(tgt["area"], 1.0) >= YELSURE else MAXDOWN
    dpitch = _cap(max(min(_deg(tgt["y"] - CY), tdown - pitch), MAXUP - pitch))
    dist = tgt["dist"]
    if dist <= reach + 0.5:
        memo["inreach_t"] = ticks          # the clock the vanish test above reads
    aimed = abs(dyaw) < 10 and abs(dpitch) < 10
    ...
    acts.append({"action": {"camera": [dpitch, dyaw]}, "repeat": 1})
    if not aimed:
        # SWING WHILE TURNING (step 510). 240 ticks bought 4 swings ... The melee hitbox is
        # wider than 10 deg and a swing at nothing costs nothing
        if abs(dyaw) < 25 and abs(dpitch) < 25 and ticks - int(memo.get("last_swing", -99)) >= SWING - 1:
            acts.append({"action": {"attack": 1}, "repeat": 1})
    ...
    if ticks - int(memo.get("last_swing", -99)) >= SWING - 1:
        acts.append({"action": {"attack": 1, "forward": 1 if (far and not under) else 0,
                                "sneak": 1 if edge else 0}, "repeat": 1})

```


\end{myskillbox}

\clearpage
%
%


\section{Harness and Implementation Details}
\label{sec:impl-details}

\lstdefinestyle{jsonrow}{
  basicstyle=\ttfamily\scriptsize,
  breaklines=true,
  breakatwhitespace=false,
  columns=fullflexible,
  keepspaces=true,
  showstringspaces=false,
  aboveskip=2pt,
  belowskip=2pt
}

This appendix covers five things:
the episode index (\Cref{app:impl-index}),
the spatial grid (\Cref{app:impl-grid}),
the configuration of each method and the action interface shared by all methods (\Cref{app:impl-configs}),
and storage and cost statistics (\Cref{app:impl-stats}).

\subsection{Episode Index}
\label{app:impl-index}

The harness appends one row to \texttt{index.jsonl} for each completed episode (\Cref{box:index-example}).
Each row contains:
\begin{itemize}
  \item the episode identifier and the path to the episode's directory;
  \item the first and last environment steps of the episode;
  \item the kind of action executed;
  \item a one-line summary;
  \item a list of tags;
  \item the agent's position $(x, y, z)$. 
\end{itemize}
The summary records the procedure call with its arguments, followed by its outcome: how long it ran, how far the agent moved, how the inventory changed, and any milestone reached.
Tags flag notable events such as a milestone, an inventory change, or a failure to move (\texttt{stuck}).
The harness computes all of these fields from the structured game state; the model does not write them.

\begin{promptbox}[label={box:index-example}]{Example rows from \texttt{index.jsonl} (one Minecraft run).}

\begin{lstlisting}[style=jsonrow]
{"action_kind": "use_at", "first_step": 2318, "last_step": 2370, "id": "ep_000043", "kind": "episode", "path": "episodes/ep_000043", "summary": "[ACTION] use_at(x=3168, y=64, z=1420, item=water_bucket, face=below, hold=1) -> 53 steps: moved 0.0, inv {'bucket': 1, 'water_bucket': -1}, milestone ['log_1']", "tags": ["milestone", "inventory_changed", "note"], "x": 3168.66, "y": 64.0, "z": 1419.47}
{"action_kind": "go_to", "first_step": 2444, "last_step": 2703, "id": "ep_000048", "kind": "episode", "path": "episodes/ep_000048", "summary": "[ACTION] go_to(x=3153.4, z=1401.4, within=0.8, n=260) -> 260 steps: moved 0.2", "tags": ["stuck", "note"], "x": 3158.30, "y": 57.0, "z": 1400.53}
\end{lstlisting}
\end{promptbox}

\subsection{Spatial Grid}
\label{app:impl-grid}

The harness also maintains \texttt{grid.jsonl}, an append-only record of the world cells the agent has interacted with.
Each row gives a cell's integer coordinates, the step at which it was recorded, and one of five kinds:
\begin{itemize}
  \item \texttt{walked}: the agent stood in the cell;
  \item \texttt{placed}: the agent placed an item in the cell;
  \item \texttt{opened}: the agent opened the screen of a block in the cell;
  \item \texttt{blocked}: the agent could not move into the cell;
  \item \texttt{hazard}: the agent took damage in the cell.
\end{itemize}
Rows of kind \texttt{placed} and \texttt{opened} also record the item and the agent's pose at that moment.
Rows of kind \texttt{hazard} record the amount and type of damage.
Below we give one example of each kind.

\begin{tcolorbox}[
    enhanced, breakable,
    colback=gray!3, colframe=gray!50,
    boxrule=0.5pt, arc=2mm,
    left=3mm, right=3mm, top=2mm, bottom=2mm,
    title={\small Example rows from \texttt{grid.jsonl}, one for each cell kind.},
    coltitle=black, colbacktitle=gray!15
]
\label{box:grid-example}
\begin{lstlisting}[style=jsonrow]
{"cell": [-3010, 71, -5573], "kind": "walked", "step": 1}
{"cell": [-3017, 49, -5569], "item": "furnace", "kind": "placed", "pose": {"pitch": 55.0, "x": -3017.7, "y": 49.0, "yaw": 270.0, "z": -5568.37}, "step": 4841}
{"cell": [-3009, 69, -5566], "item": "crafting_table", "kind": "opened", "pose": {"pitch": 55.0, "x": -3007.43, "y": 69.0, "yaw": 90.0, "z": -5565.54}, "step": 1206}
{"cell": [-3008, 61, -5569], "kind": "blocked", "step": 1961}
{"cell": [-3019, 34, -5569], "damage": 3.0, "kind": "hazard", "step": 6361, "what": "damage"}
\end{lstlisting}
\end{tcolorbox}

\subsection{Method Configurations}
\label{app:impl-configs}

\Cref{tab:arm-capabilities} lists the configuration of each method on Minecraft. All methods receive the same 12\,KB standing brief in prompt.
The atoms are closed-loop: they read the structured game state at every step.
No method has access to task-specific macros such as tree harvesting, strip mining, or portal building.

We reimplemented PRO-LONG and VISTA in our harness without changing their core algorithms.
PRO-LONG keeps its programmatic text logs, and VISTA keeps its archive of visual observations and its inspect interface.

\textbf{Matched inspection, different addressing.} VISTA and MFM have equivalent visual inspection capabilities: both store every observation losslessly with its step identity, and both let the agent reopen earlier frames, crop and enlarge regions, and compare several past or current views in one context. The methods differ only in how the agent locates evidence and what it does with it. VISTA addresses frames by turn and frame identifier; MFM additionally provides a harness-written index of cues (time, position, action type, tags, outcome summary) that the agent can query, and agent-written memory that cites the supporting episodes. Our comparison with VISTA therefore isolates the contribution of the addressing layer and consolidation, not of the inspection tools.

\textbf{Why only MFM writes executable procedures.} Neither baseline provides a persistent place for agent-written procedures: PRO-LONG's durable memory is a single harness-owned log, and VISTA's writable memory is limited to two notes files. Turning an action sequence into a procedure also requires checking whether that sequence has succeeded before. MFM's associative index makes this check direct: the agent can query past episodes by action type and outcome tags, then inspect the linked frames to confirm the result. VISTA addresses frames only by turn identifier, and PRO-LONG's text log, while searchable, has no structured action or outcome fields and no links to visual evidence, so the same check requires scanning and reconstructing past attempts. Without these supports, the baselines kept their procedural knowledge as prose notes rather than callable procedures.

\begin{table}[t]
\centering
\small
\caption{\textbf{Configuration of each method on Minecraft tasks.}
All methods share the action and procedure reference, the knowledge sections, the ground-truth rules, and the per-turn state block; Direct Evaluation omits the block's \texttt{[RECENT]} and \texttt{[MAP]} fields.
All methods with memory can write files and execute stored procedures through the same interface.
}
\label{tab:arm-capabilities}
\setlength{\tabcolsep}{4pt}
\renewcommand{\arraystretch}{1.15}
\begin{tabularx}{\textwidth}{@{}>{\raggedright\arraybackslash}p{3.1cm} >{\raggedright\arraybackslash}X >{\raggedright\arraybackslash}X >{\raggedright\arraybackslash}X >{\raggedright\arraybackslash}X@{}}
\toprule
\textbf{Capability} & \textbf{Direct} & \textbf{PRO-LONG} & \textbf{VISTA} & \textbf{MFM} \\
\midrule
Standing brief & 12\,KB & 12\,KB & 12\,KB & 12\,KB 
\\
Per-turn \texttt{[RECENT]} / \texttt{[MAP]} & no (as direct agent is not supposed to have any memory) & yes & yes & yes \\
Episode store and query tool & no & no 
& archived frames via \texttt{inspect} & yes \\
Write access to own files; procedure execution & yes & yes & yes & yes \\
Agent-written procedures (observed) & 0 & 0 & 0 & 4 written / 14 calls \\
Plan cap & 400 steps & 400 steps & 400 steps & 400 steps \\
Images per call & current frame & current frame & current frame + inspected frames & current frame + inspected frames \\
\bottomrule
\end{tabularx}

\end{table}

\subsection{Storage and Cost}
\label{app:impl-stats}

All main and ablation results are averaged over 3 seeds; each seed fixes the Minecraft world or KiCad task order, with further variation from stochastic model sampling. In total we ran 96 Minecraft runs (72 main + 24 ablation, capped at 12k steps for Diamond and 36k for Dragon) and 1,125 KiCad episodes ($25\text{ tasks} \times 15\text{ configurations} \times 3\text{ seeds}$). All  48 Dragon runs exhaust their budget
(Fig.~\ref{fig:steps-per-turn}), consuming $\approx$1.7M steps and up to 14 hours each. In the Claude Opus 5 on Dragon configuration, a single run makes $0.9\text{--}3.9\text{k}$ model calls and costs $\$108\text{--}\$969$ (Table~\ref{tab:dragon-opus-cost-36k}). 

\Cref{tab:mfm-numbers} reports storage, retrieval, and memory-use statistics for one MFM run.
\Cref{tab:dragon-opus-cost-36k} reports the cost of one Kill the Ender Dragon run per method with Claude Opus~5. 
Both tables come from single runs and are meant to show orders of magnitude.

\begin{table}[ht]
\centering
\small
\caption{\textbf{Storage, retrieval, and memory-use statistics} for the GPT-6-Astra Mine Diamond run that reached all 10 milestones.
}
\label{tab:mfm-numbers}
\renewcommand{\arraystretch}{1.1}
\begin{tabular}{@{}lr@{}}
\toprule
\textbf{Quantity} & \textbf{Value} \\
\midrule
Environment steps & 12,000 \\
Model calls & 130 \\
Episodes stored & 636 \\
Episode directory on disk & 46\,MB \\
Index (\texttt{index.jsonl}) size & 313\,KB \\
Full-resolution frames on disk & 720\,MB \\
Query time over the full index & 0.3--0.4\,s \\
\texttt{view\_image} calls & 87 \\
Agent-written procedures (written / calls) & 4 / 14 \\
\bottomrule
\end{tabular}
\end{table}

\begin{table*}[ht]\centering\small
\caption{Costs of Ender Dragon runs with Claude Opus 5 (36,000 steps) using Claude Code. 
}
\label{tab:dragon-opus-cost-36k}
\setlength{\tabcolsep}{4pt}
\begin{tabular}{@{}llrrrrrrl@{}}\toprule
\textbf{Method}   & \textbf{Milest. obs.}  & \textbf{Calls} & \textbf{Tokens (M)} & \textbf{USD (\$)} & \\ \midrule
MFM   & 10  & 2,129 & 47 & 108  \\
VISTA    & 9 & 2,177 & 69 & 172  \\
PRO-LONG    & 8  & 895 & 57 & 168  \\
Direct   & 6 & 3,934 & 155 & 969  \\
\bottomrule\end{tabular}
\end{table*}

\clearpage
\section{Benchmark Setup and Prompts}
\subsection{MCU Tasks Descriptions}
\label{sec:task-details}
\textbf{Mine Diamond from Scratch.} The agent starts with no resources and must progress through the Minecraft technology tree: gathering wood, stone, and iron, crafting increasingly capable tools, and ultimately mining a diamond.

\textbf{Kill Ender Dragon.} The agent completes the same early-game progression, then prepares diamond equipment, enters the Nether to obtain blaze materials, crafts Eyes of Ender, reaches the End, and defeats the Ender Dragon.

\begin{figure*}[ht]
  \centering

    \centering
    \includegraphics[width=\linewidth]{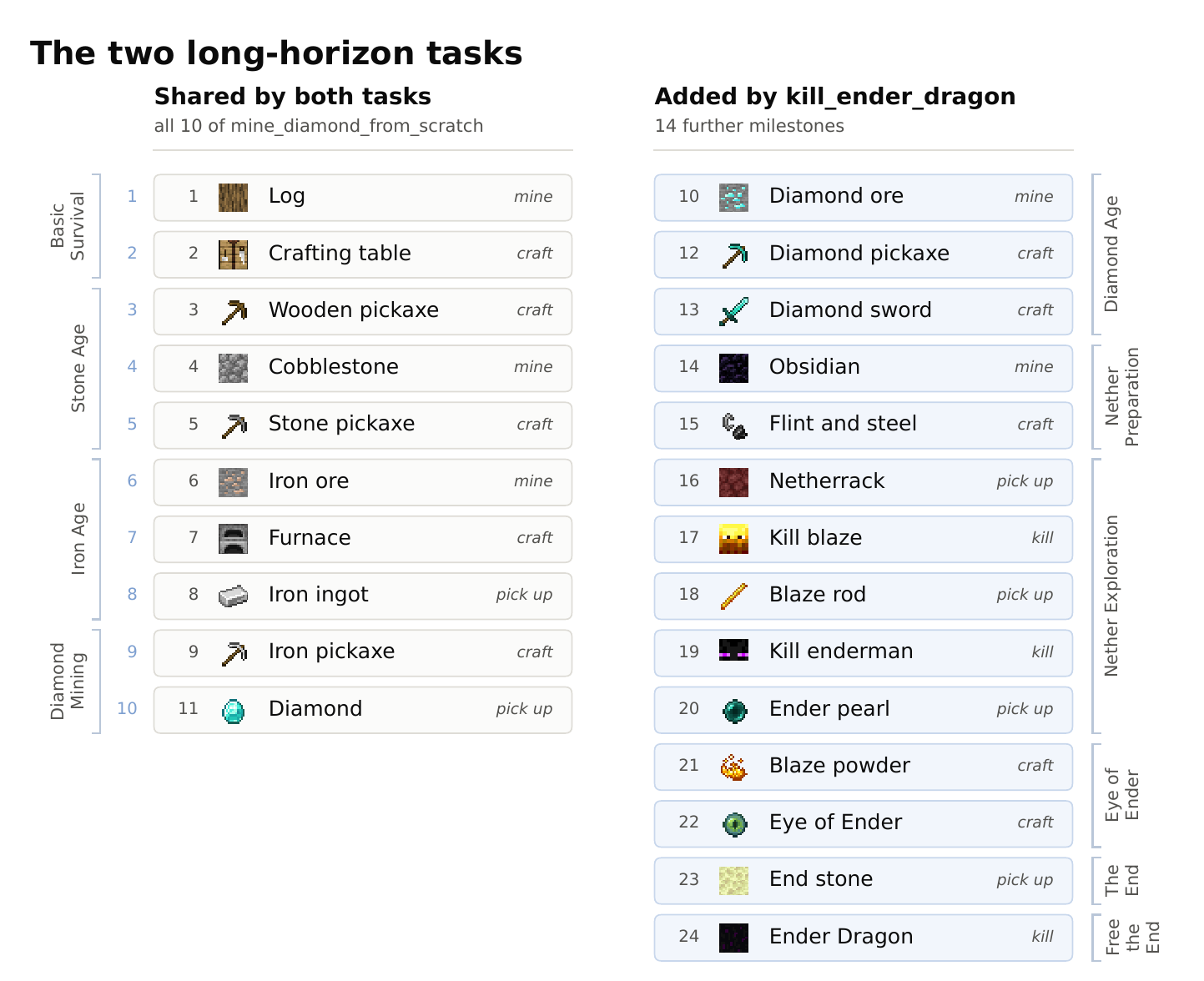} 

  \caption{\small \textbf{Milestone progression for the Minecraft tasks.} Mine Diamond from Scratch contains 10 milestones, while Kill Ender Dragon shares the same initial progression and adds 14 milestones required to reach and defeat the Ender Dragon. Numbers in the left column index the Mine Diamond list; inner and right-column numbers index the Ender Dragon list, in which Diamond ore is inserted as milestone 10, so Diamond becomes 11 and the added milestones continue from 12. \textbf{Note that the milestones do not need to be finished strictly in this order.} }
  \label{fig:task-milestones}
\end{figure*}

\subsection{New Tasks for Evaluating Skill Reuse}
We design three additional Minecraft tasks to study whether reusing
agent-written skills reduces the steps required to solve new tasks.
These tasks introduce new objectives while providing opportunities
to apply procedures acquired during earlier interaction.

\textbf{Iron Shield.} The agent builds a furnace, mines iron ore,
smelts an iron ingot, and crafts a shield, combining intermediate
objectives familiar from Mine Diamond from Scratch with a new crafting
goal. A small vein of iron ore is placed underground, 13 blocks below
and 12 blocks south of the starting position, at coordinates given in
the task description.

\textbf{Climb Out.} The agent starts sealed in a two-block pocket deep
underground and must return to the surface, 40 blocks above, and mine a
block of bricks set in the ground there; the block's coordinates are
given in the task description. Reaching it requires digging upward,
whereas the example procedures supplied to the agent only dig downward
or level.

\textbf{Cook Meat.} The agent builds a furnace, kills an animal, and
cooks its meat, introducing a hunting objective absent from Mine
Diamond from Scratch. Three cows are placed beside the starting
position.
In all three tasks, early-game resources are supplied (a stone pickaxe
and cobblestone, plus a crafting table and logs for Iron Shield and
Cook Meat, and a stone sword for Cook Meat) and the target is either
placed nearby or located in the task description, reducing the need to
repeat initial resource gathering and search.

We compare the steps required to complete each task with and without
access to previously written skills. This evaluation tests whether
procedures learned from earlier experience improve efficiency when
applied to new objectives.
\label{sec:}
\begin{figure*}[ht]
    \centering
        \centering
        \includegraphics[height=4.2cm,width=\linewidth,keepaspectratio]
            {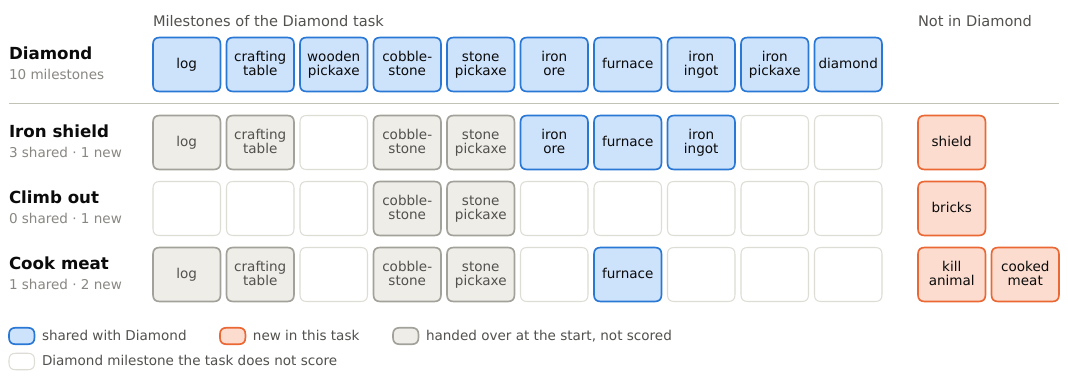}
 \caption{\small \textbf{Milestone structure of Diamond and three related tasks.} Columns align the tasks by Diamond’s ten milestones. Blue indicates shared, scored milestones; orange indicates task-specific milestones; gray indicates milestones provided at initialization and excluded from scoring. Empty cells denote Diamond milestones not scored in the corresponding task.
    }
    \vskip -0.2cm
\end{figure*}

\subsection{Prompts and Outputs Examples}
\label{sec:prompts}
In this section, we provide examples of prompts used.

\begin{promptbox}[label={box:nether}]{Prompts of the Nether Portal case study}
Build a nether portal and step through it. You are standing on a cleared stone pad and you already carry everything you need, with plenty to spare: 64 obsidian, 64 cobblestone, a flint and steel and a diamond pickaxe. The frame is a vertical rectangle 4 wide and 5 tall lying in one plane, either north-south or east-west. Its interior is
2 blocks wide and 3 blocks tall, and every one of those 6 interior cells must be empty
air. The 10 blocks that matter are 2 below the interior, 2 above it, and 3 up each side; the 4 corners of the rectangle are not part of the portal and may be anything or nothing. Then strike the flint and steel inside the frame, on top of one of the two, and the interior turns purple;
walk into it and wait to arrive in the
Nether. The only thing this task asks of you is to get the shape right. You have far more obsidian than a portal needs, the cobblestone is there to stand on so you can reach the upper rows, and the diamond pickaxe will take back any block you want to move.
\end{promptbox}

\begin{promptbox}[label={box:memory-paragraph}]{The Memory paragraph of the standing brief.}

**Memory.** `./logs.txt` and `episodes/` are the complete record of this life
(\texttt{./tools/episodes.py -{}-last N | -{}-steps A B | -{}-near X Z | -{}-tag T | -{}-grep text
[-{}-sheet out.png] [-{}-then-now <id>] -{}-q "<question>"}); 
\texttt{mine/} is yours
and is carried to your next life, and \texttt{mine/working.md} is put in front of you every turn. The log knows where; the frame
knows what: \texttt{view\_image} a frame before you walk to a remembered place and, on arrival,
\texttt{./tools/episodes.py -{}-then-now <that episode> --q "..."} once; look after a hit, and when a plan's outcome surprised you.

\end{promptbox}



\begin{promptbox}[label={box:consolidation}]{Consolidation prompt.}

**Working memory.** `mine/working.md` is yours and is put in front of you every turn
as `[WORKING]`: the plan you are on and why, its next visible prediction, the
contradictions you have not resolved, and the entries or frames you would re-open
(paths). It survives a context reset, so the turn after one starts from it instead of
from the log. Rewrite it when the plan changes, not every turn; keep it under 40
lines. Your digest or `GUIDE.md` under `mine/` is what is true of this world across
lives; `working.md` is what you are doing now.

\end{promptbox}

\begin{promptbox}[label={box:per-turn}]{Minecraft: per-turn observation block.}

Step 8/36000 -- 35992 steps remain. Environment-verified progress: 1/24 milestones. Score = COUNT achieved, in ANY order -- bank whichever are already within reach (a furnace needs only 8 cobblestone) before chasing the next in list order. First unmet in list order: `crafting\_table` (craft\_item of crafting\_table).

Act turns since anything verifiably changed (milestone or inventory): 0.

Milestones:

  [x] mine\_log (step 8)
  
  [ ] crafting\_table 
  
  [ ] wooden\_pickaxe 
  
  
  
  
  
  
  
  
  
  
  
  
  
  
  
  
  
  
  
  
  

[STATE] pos=(3.4, 64.0, 2.0) pitch=0 yaw=0

Vitals: health=20/20 food=20/20

Inventory (35 slots free): oak\_logx2

Hotbar: hotbar.1=oak\_logx2

Held: unknown

Nearby entities: none reported

GUI open: False

[CURSOR] no screen open

[RECENT] the last entries that finished, from the harness's record:

- step 5 {"forward":1} x5: moved 0.0

- step 8 {"attack":1} x3: moved 0.9, inv oak\_log +1, events MILESTONE mine\_log verified; mine\_block oak\_log +1

[MAP] what the runner knows for certain from this life's own positions and events (exact unless marked a sighting):

- You stand in cell (3, 64, 2); 3 cells walked this life (a route over them is walkable again).

\end{promptbox}

\begin{promptbox}[label={box:}]{KiCad: action reference.}

Coordinates are normalized to [0, 1000] on both axes: (0,0) is the top-left of the screen, (1000,1000) the bottom-right — always fractions of the screen times 1000, NEVER raw pixels.

Available actions (each an object in an "actions" list):

  {"action": "click", "coordinate": [x, y], "button": "left|right|middle", "clicks": 1|2|3}
  
  {"action": "mouse\_move", "coordinate": [x, y]}
  
  {"action": "drag", "start\_coordinate": [x, y], "coordinate": [x, y], "button": "left"}
  
  {"action": "scroll", "direction": "up|down|left|right", "amount": N, "coordinate": [x, y]}
  
  {"action": "type", "text": "...", "press\_enter": false}      
  
  {"action": "key", "keys": ["ctrl", "s"]}                     
  
  {"action": "key\_down", "keys": [...]} / {"action": "key\_up", "keys": [...]}
  
  {"action": "hold\_key", "keys": [...], "duration": seconds}
  
  {"action": "wait", "duration": seconds}
  
  {"action": "screenshot"}      

\end{promptbox}

\begin{promptbox}[label={box:}]{KiCad: task framing and rules.}

You are operating a real Linux desktop through a GUI to complete a task. The desktop is XFCE with KiCad installed; work inside KiCad's own editors (schematic editor etc.) using its menus and keyboard shortcuts (e.g. 'a' to add a symbol, 'w' to wire, ctrl+s to save). The evaluator checks real system state at the end -- it compares the saved project's netlist against a reference -- so save the project before terminating.

**Rules that have cost episodes before**:
- Before typing into any app or field, verify in the CURRENT screenshot that the target window/field exists and is focused -- text typed blind goes to whatever is active and is lost.
  
- Never batch input in the same turn that opens the app or dialog it targets; open, look, then act.

- After typing something long, check the screenshot that it appears complete before
  pressing enter.
  
- Save the project (ctrl+s, confirm any dialog) and verify the title bar no longer shows
  an unsaved marker before terminating. Terminate as soon as the task is durably done;
  every further step is budget, not progress.
  
- If the screen has not changed after the same action twice, the action is not doing
  what you think: zoom in, re-read the log, and change approach.

\end{promptbox}

\end{document}